%% file: fullpaper_Final.tex
\documentclass[10pt]{article}

\usepackage{spconf_modified,amsmath,epsfig}

\usepackage{epsfig,cite}
\usepackage{amsmath}
\usepackage{amssymb}
\usepackage{color}
\usepackage{url}
\usepackage{multirow}
\usepackage{graphicx}
\usepackage{epstopdf}
\newcommand{\blue}[1]{\textcolor[rgb]{0,0,0}{#1}}   
\newcommand{\white}[1]{\textcolor[rgb]{1,1,1}{#1}}  

\renewcommand{\baselinestretch}{1}

\title{Advances in Hyperspectral Image Classification\\
{\small [Earth monitoring with statistical learning methods]}}
\name{Gustavo Camps-Valls, Devis Tuia, Lorenzo Bruzzone and J{\'o}n Atli Benediktsson}
\address{}

\begin{document}
\maketitle


\input{01_introduction.tex}
\input{03_advanced.tex}

\input{04_spatial.tex}
\input{05_adaptation.tex}

\input{06_conclusions.tex}

\section*{\small ACKNOWLEDGMENTS}

\blue{This work was partially supported by the Spanish Ministry of Economy and Competitiveness (MINECO) under project LIFE-VISION TIN2012-38102-C03-01, and by the Swiss National Research Foundation under grant PZ00P2-136827.}

\section*{\small AUTHORS}

\blue{
\noindent {\bf\em Gustavo Camps-Valls} (gustavo.camps@uv.es) (SM'07) received a Ph.D. degree in Physics (2002) from the Universitat de Val\`encia, Spain, where he is currently an Associate Professor in the Electrical Engineering Dept. and head of the Image and Signal Processing group. His research is tied to machine learning for signal and image processing, with special focus on remote sensing data analysis. He has co-authored many journal papers and edited several books on kernel methods. The ISI lists him as a highly cited researcher. He is member of the Machine Learning for Signal Processing TC of the IEEE SPS, and serves as Associate Editor for IEEE Transactions on Signal Processing, IEEE Signal Processing Letters, and IEEE Geoscience and Remote Sensing Letters. Visit http://isp.uv.es/ for more information.\\
\noindent {\bf\em Devis Tuia} (devis.tuia@epfl.ch) received a Ph.D in environmental sciences at the Univeristy of Lausanne (Switzerland) in 2009. He was  then a postdoc researcher at both the University of Val\`encia, Spain and the University of Colorado at Boulder. He is now senior research associate at the laboratory of Geographic Information Systems, at the Ecole Polytechnique F{\'e}d{\'e}rale de Lausanne (EPFL). His research interests include the development of algorithms for information extraction and classification of very high resolution remote sensing images using machine learning algorithms. Visit http://devis.tuia.googlepages.com/ for more information.\\
\noindent {\bf\em Lorenzo Bruzzone} (lorenzo.bruzzone@ing.unitn.it) received the Ph.D. degree in Telecommunications Engineering from the University of Genoa, Italy, in 1998. He is currently a Full Professor at the University of Trento, Italy where he is the founder and the head of the Remote Sensing Laboratory. His current research interests are in the areas of remote sensing, radar and SAR, signal processing, and pattern recognition. He is the author (or coauthor) of many highly cited scientific publications in referred international journals, conference proceedings, and book chapters. He is the co-founder of the IEEE International Workshop on the Analysis of Multi-Temporal Remote-Sensing Images (MultiTemp) series. Since 2003 he has been the Chair of the SPIE Conference on Image and Signal Processing for Remote Sensing. Since 2013 he has been the founder Editor-in-Chief of the IEEE Geoscience and Remote Sensing Magazine. He is a member of the Administrative Committee of the IEEE Geoscience and Remote Sensing Society (GRSS) and a Distinguished Speaker of the IEEE GRSS. He is Fellow of IEEE. Visit http://rslab.disi.unitn.it for more information.\\
\noindent {\bf\em J{\'o}n Atli Benediktsson} (benedikt@hi.is) received the Ph.D. degree in electrical engineering from Purdue University, West Lafayette, IN in 1990. He is currently Pro Rector for Academic Affairs and Professor of Electrical and Computer Engineering at the University of Iceland. His research interests are in remote sensing, biomedical analysis of signals, pattern recognition, image processing, and signal processing. He was a co-founder of Oxymap, Inc. Prof. Benediktsson was the 2011-2012 President of the IEEE Geoscience and and Remote Sensing Society. He was Editor-in-Chief of the IEEE Transactions on Geoscience and Remote Sensing (2003-2008). He is a Fellow of IEEE and a Fellow of SPIE. Visit www.hi.is/$\sim$benedikt for more information.
}

\section*{\small REFERENCES}

\footnotesize{
\input{fullpaper_Final.bbl}}

\end{document}

%% file: 01_introduction.tex
\section*{\small{INTRODUCTION}} \label{sec:intro}

The technological evolution of optical sensors over the last few decades has provided remote sensing analysts with rich spatial, spectral, and temporal information. In particular, the increase in spectral resolution of hyperspectral images and infrared sounders opens the doors to new application domains and poses new methodological challenges in data analysis. Hyperspectral images (HSI) allow to characterize the objects of interest (for example land-cover classes) with unprecedented accuracy, and to keep inventories up-to-date. Improvements in spectral resolution have called for advances in signal processing and exploitation algorithms. This paper focuses on the challenging problem of hyperspectral image classification, 
which has recently gained in popularity and  attracted the interest of other scientific disciplines such as machine learning, image processing and computer vision. In the remote sensing community, the term `classification' is used to denote the process that assigns single pixels to a set of classes, while the term `segmentation' is used for methods aggregating pixels into objects, then assigned to a class. 

One should question, however, what makes hyperspectral images so distinctive. Statistically, hyperspectral images are not extremely different from natural grayscale and color photographic images~(see chapter 2 of~\cite{CampsValls11mc}). Grayscale images are spatially {\em smooth}: the joint probability density function (PDF) of the luminance samples is highly uniform, the covariance matrix is highly non-diagonal, the autocorrelation functions are broad and have generally a $1/f$ band-limited spectrum. In the case of color images, the correlation between the tristimulus values of the natural colors is typically high. While the three tristimulus channels are equally smooth in generic RGB representations, opponent representations imply an uneven distribution of bandwidth between channels. Despite all these commonalities, the analysis of hyperspectral images turns out to be more difficult, especially because of the high dimensionality of the pixels, the particular noise and uncertainty sources observed, the high spatial and spectral redundancy, and their potential non-linear nature. Such nonlinearities can be related to a plethora of factors, including the multi-scattering in the acquisition process, the heterogeneities at subpixel level, as well as the impact of atmospheric and geometric distortions. These characteristics of the imaging process lead to distinct nonlinear feature relations, i.e. pixels lie in high dimensional complex manifolds. The high spectral sampling of HSI (of the order of one band each 5-10 $nm$ in the electromagnetic spectrum) also leads to strong collinearity issues. Finally, the spatial variability of the spectral signature increases the internal class variability. All these factors, in conjunction to the few labeled examples typically available, make HSI image classification a very challenging problem. As a result, the accuracy obtained with standard parametric classifiers commonly used for multispectral image classification is typically compromised when applied to HSI~\cite{CampsValls05}. 

Many of these limitations have been recently addressed under the framework of statistical learning theory (SLT)~\cite{Vapnik98}. SLT is a general framework for learning functions from data, which reduces to finding a \blue{linear function defined in a high (eventually infinite) dimensional Hilbert feature space $f\in{\mathcal H}$ that learns the relation between observed input-output data pairs $(x_1,y_1),$ $\ldots,$ $(x_\ell,y_\ell)\in{\mathcal X}\times{\mathcal Y}$, and that {\em generalizes} well.} 
Generalization is the capability of a method to extrapolate to unseen situations, i.e. the function $f$ should accurately predict the label $y^\ast\in{\mathcal Y}$ for a new input example $x^\ast\in{\mathcal X}$. 
Generalization has recurrently appeared in  statistics literature for decades under the names of bias-variance dilemma, capacity control, or complexity regularization trade-off. The underlying idea is to constrain too flexible functions in order to avoid overfitting the training data. 

The SLT framework formalizes this intuition~\cite{Vapnik98} and seeks for prediction functions that optimize a \blue{functional ${\mathcal L}_{reg}$ that takes into account {\em both} an empirical estimation of the training error (loss), ${\mathcal L}_{emp}$, 
and an estimate of the complexity of the model (or regularizer), $\Omega(f)$:
$$
\begin{array}{lll}
{\mathcal L}_{reg} &=& {\mathcal L}_{emp} + \lambda~\Omega(f) \\[2mm]
& = &  \sum_{i=1}^\ell V(x_i,y_i,f(x_i)) + \lambda~\Omega(f) 
\end{array}
$$
where $V$ is a loss function acting on the $\ell$ labeled samples, and $\lambda$ is a trade-off parameter between the cost and the regularization.}
Different losses and regularizers can be adopted for solving the problem, involving completely different families of models and solutions. 
\blue{To ensure unique solutions, many SLT algorithms use strictly convex loss functions. The regularizer $\Omega(f)$ limits the capacity of the classifier to minimize ${\mathcal L}_{emp}$ and favors smooth functions. } 

The hyperspectral image processing community has contributed to the design of specific loss functions and regularizers to take the most out of the acquired images. For example, regularization appears explicitly in many HSI classifiers when trying to impose the spatial homogeneity of images, when including the wealth of user's labeling in active learning, or when exploiting the information contained in the unlabeled pixels to better describe the image manifold in semisupervised learning. 
Classifiers should also be robust to changes in the image representation: small perturbations of pixels and objects in the image manifold should not produce big differences in the classification. This is why the inclusion of proper image representations and invariances is also an active field.

In this paper, we review the recent advances in HSI classification under the SLT framework. Section II presents  advances in HSI classification using active, semisupervised and sparse learning approaches. Section III summarizes the field of spatial-spectral regularization both in terms of feature extraction and advanced classifiers. Section IV covers the  field of adaptation of both classifiers and feature representations, and reviews solutions to encode invariances in HSI classifiers. We conclude in Section V and outline future challenges.

%% file: 03_advanced.tex
\section*{\small{ADVANCED REGULARIZED IMAGE CLASSIFICATION}}\label{sec:advanced}


Before HSI, most of the classifiers used in remote sensing were parametric, such as Gaussian maximum likelihood or linear discriminant analysis. These methods, based on the estimate of the covariance matrix, were successful when dealing with early multispectral images, whose dimensionality was usually comprised between four and ten bands. HSI changed the rules, as the increased dimensionality of pixels raised to hundreds. Standard parametric methods became either unfeasible or unreliable, since estimating the class-covariance matrices requires many labeled samples, which are usually not available. 
For that reason, research turned to include regularization, 
either explicitly through Tikhonov's terms in the involved covariance matrices, or by perform classification in a subspace of reduced dimensionality~\cite{Bandos09,Jensen10}. 

Although successful, parametric models make strong assumptions about the normality of the class conditional PDFs or about the linearity of the problem. Due to the complexity of HSI, such assumptions rarely hold, which encouraged research towards non-parametric and nonlinear models. 
\blue{Several approaches have been introduced in the last decade in the field of hyperspectral image classification: kernel methods and support vector machines (SVM)~\cite{CampsValls05}, sparse multinomial logistic regression~\cite{Li12}, {neural networks~\cite{Ratle10}} and Bayesian approaches like relevance vector machines~\cite{Mianji11} and Gaussian Processses classification~\cite{Bazi10}. Nevertheless, the SVM has undoubtedly become the most widely used method in HSI classification research~\cite{CampsValls09}.} Unlike other nonparametric approaches, such as regularized RBF neural networks, SVM naturally implements regularization through the concept of maximum margin: \blue{given a linear classification function $f(x) = w^\top x + b$, maximizing the linear separability between classes is equivalent to minimize the $\ell_2$-norm of model weights $w$ used as regularizer, $\Omega(f)=\|w\|_2^2$.} Nonlinearity is also implemented via reproducing kernels, which allows to work in high dimensional Hilbert spaces implicitly, while still resorting to linear algebra operations~\cite{CampsValls05}.

The effectiveness of SVM rapidly turned out to be insufficient to exploit the rich information contained in HSI. In all methodologies that follow, the functional to be optimized will consider additional information 
such as the one contained in unlabeled samples, ancillary data, or distinct signal characteristics. Such heterogeneous information is typically included in the HSI classifiers through additional regularizers. 

\begin{figure}[b]
\begin{tabular}{ccc}
Supervised & After SSL & After AL \\
\includegraphics[height=2.2cm]{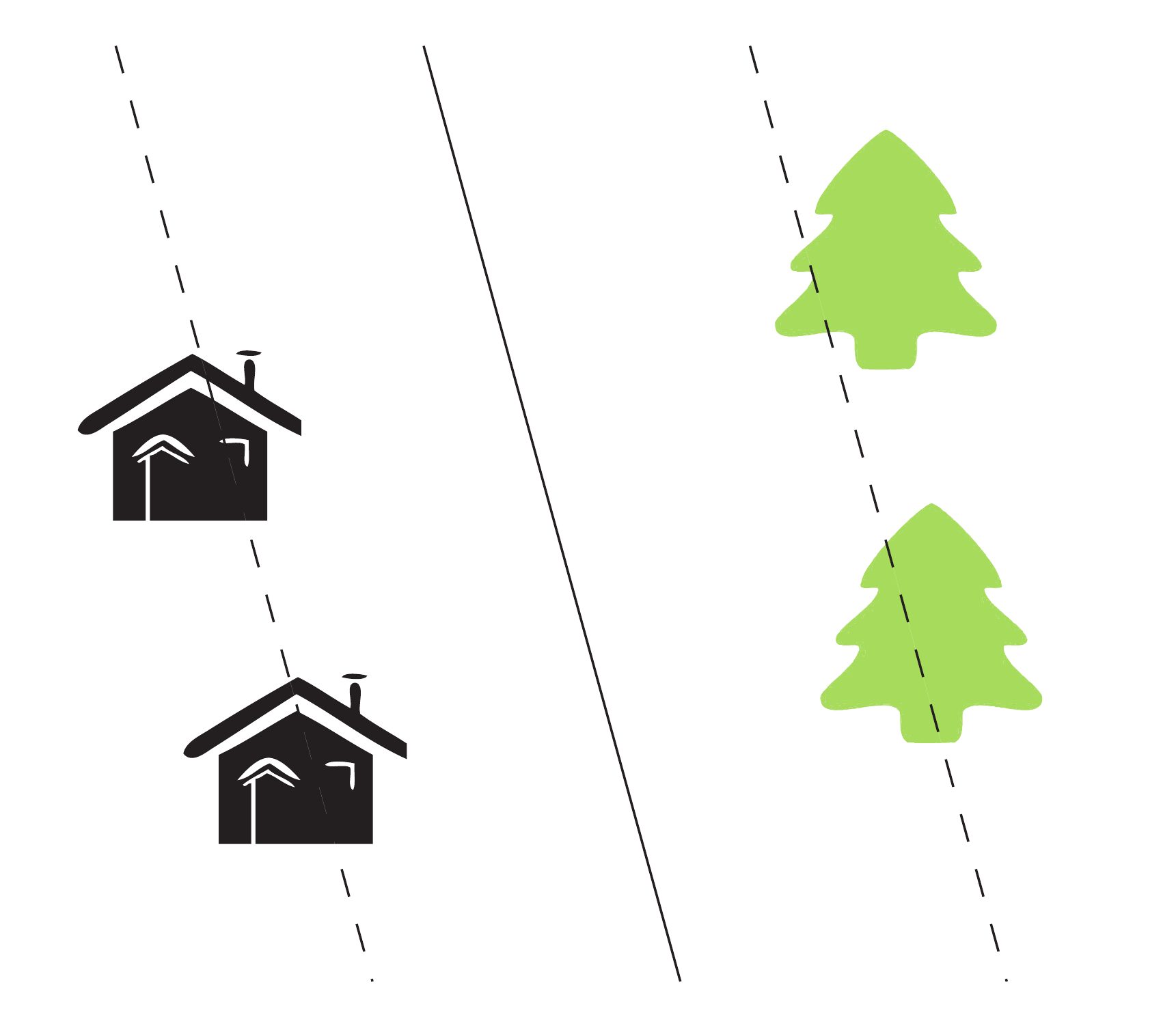}&
\includegraphics[height=2.2cm]{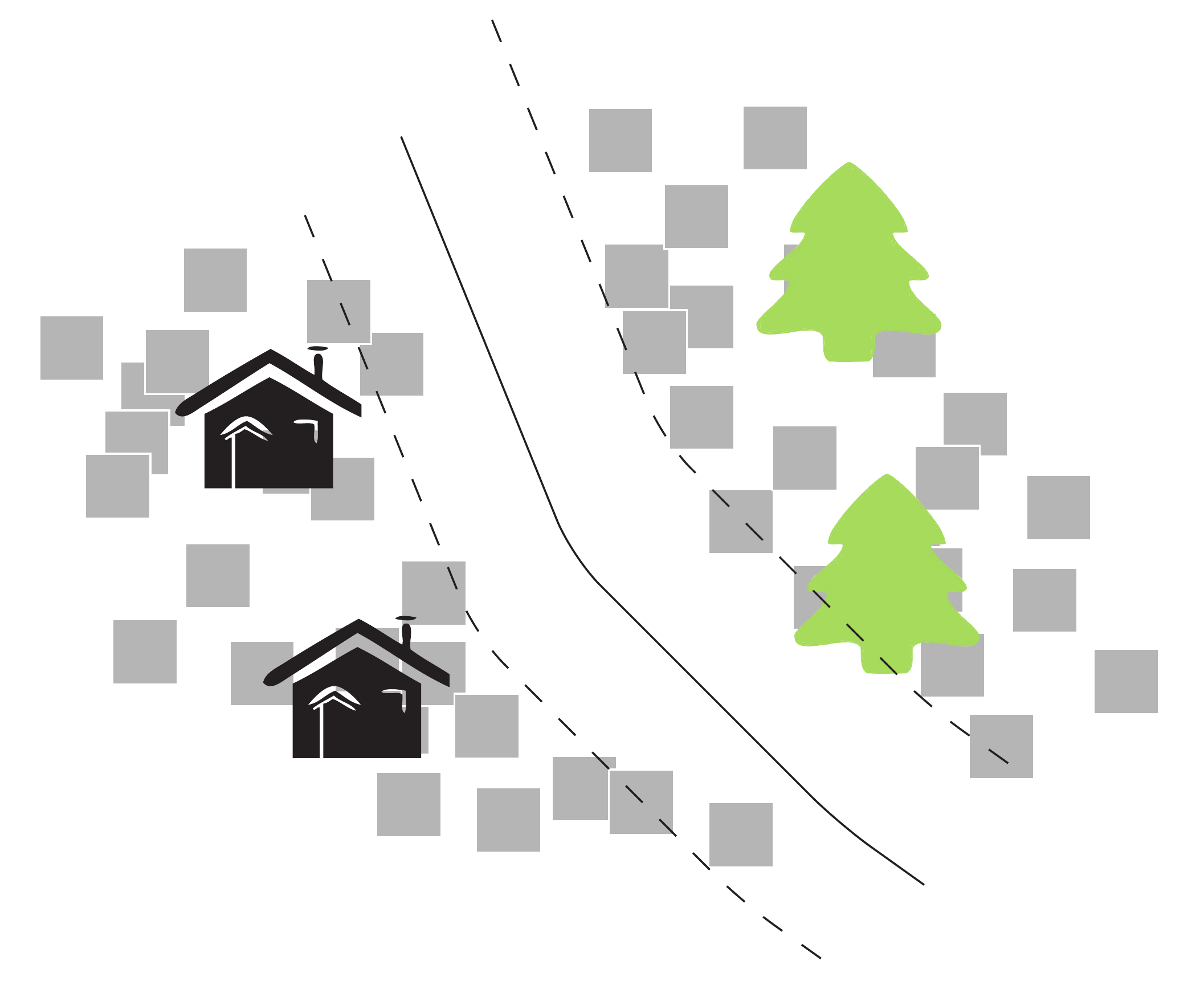}&
\includegraphics[height=2.2cm]{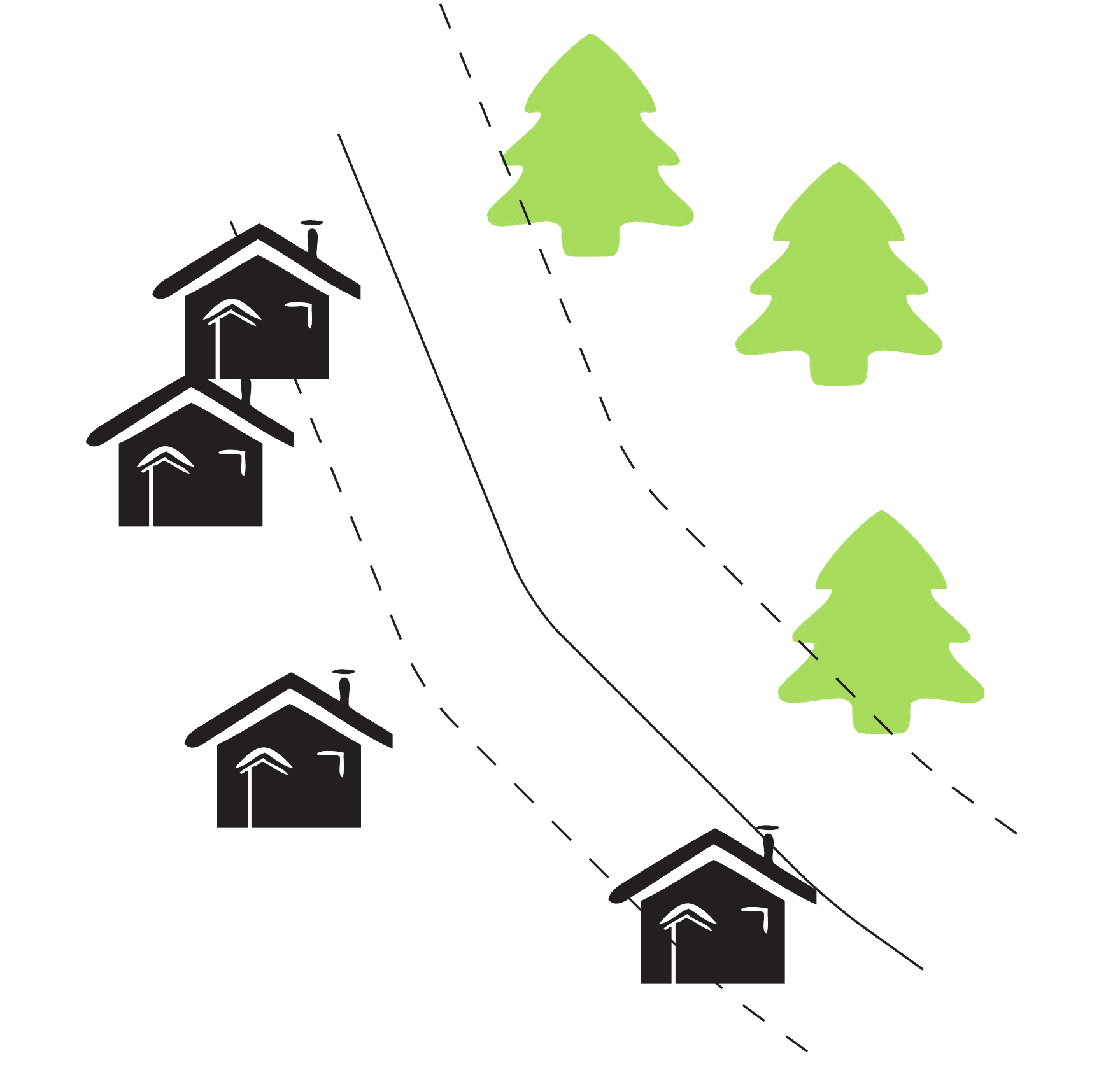}\\
\end{tabular}
\caption{Regularization of models with unlabeled samples: (left) purely supervised solution, (middle) semisupervised solution exploiting low density areas, and (right) active learning solution, where three new samples are labeled by a user.}
\label{fig:SSAL}
\end{figure}

\vspace{-0.25cm}\paragraph*{\small Regularization with unlabeled samples.} Recently, researchers started to exploit the abundant unlabeled information contained in the image itself, and new forms of regularization and priors were introduced. This is the field of \emph{semisupervised learning} (SSL, \cite{Bruzzone06,CampsValls07,Tuia09b}, central panel of Fig.~\ref{fig:SSAL}), where the minimization functional is modified to take into account 
the structure of the hyperspectral image manifold. 
Usually, semisupervised algorithms modify the decision function of the classifier by adding an extra regularization term $\Omega_u$ that acts on \blue{both labeled and unlabeled examples: 
$$
\begin{array}{ll}
{\mathcal L}_{reg} =& {\mathcal L}_{emp} + \lambda~\Omega(f) + \lambda_u~\Omega_u(f).
\end{array}
$$
Several strategies to design the regularizer have been presented. One may use the graph Laplacian 
 as a metric on the predictions to build $\Omega_u$. 
Since regularization is performed on a proximity graph, the assumption enforced is that decisions on neighboring pixels in the data manifold should be similar~\cite{Gomez08}. Another possibility considers  
regularizers that enforce wide and empty SVM margins~\cite{Bruzzone06}.} Other strategies deform the kernel function by changing the metric induced using the unlabeled samples
~\cite{Tuia09b,GomezChova10}.

\begin{table}[!t]
\begin{center}
\caption{Summary of semisupervised algorithms used in HSI classification.}
\label{tab:ssl}
\small
\setlength{\tabcolsep}{2pt}
\begin{tabular}{p{1.7cm}|p{1.5cm}|p{5cm}}
\hline
Assumption&Model &  Idea   \\\hline\hline
Low-density & TSVM~\cite{Bruzzone06} & Look for the emptiest margin \\
\hline
Manifold & Label propagation~\cite{CampsValls07} &  Spread class information on the graph of labeled/unlabeled (nearby samples are classified in the same class) \\
\cline{2-3}
& Laplacian SVM~\cite{Gomez08} & SVM hinge loss plus Laplacian eigenmaps for manifold regularization: Pixels close in the input space are also close in the graph (nearby samples are mapped close together) \\
\cline{2-3}
& SSNN~\cite{Ratle10} & Neural network trained with gradient descent replaces SVM, graph regularization with loss that forces similar pixels to be mapped closely and dissimilar ones to be separated\\
\cline{2-3}
\hline
Cluster & Cluster kernel~\cite{Tuia09b} & Increase the similarity measure (kernel) if samples fall in the same cluster, then run standard SVM \\
\cline{2-3}
& Mean map kernel~\cite{GomezChova10} & Increase similarity if samples are mapped close to centroids in Hilbert space, then run standard SVM \\\hline
\end{tabular}
\end{center}
\end{table}

We evaluate the performance of semisupervised algorithms in an AVIRIS image acquired over the Kennedy Space Center (KSC), Florida in 1996, with a total of $224$ bands of $10$~nm bandwidth with center wavelengths from 400-2500 nm. The data was acquired from an altitude of $20$ km and has a spatial resolution of $18$ m. After removing low SNR bands and water absorption bands, a total of $176$ bands remains for the analysis. The dataset originally contained $13$ classes representing the various land cover types of the environment. Many different marsh subclasses were merged in a `marsh' class, resulting into the following $10$ classes: `Water' ($761$ labeled pixels), `Mud flats' ($243$), `Marsh' ($898$), `Hardwood swamp' ($105$), `Dark/broadleaf' ($431$), `Slash pine' ($520$), `CP/Oak' ($404$), `CP Hammock' ($419$), `Willow' ($503$), and `Scrub' ($927$).  \blue{The high  dimensionality and number of classes and subclasses pose challenging problems for the classifiers, especially when few labeled examples are available.}

\begin{figure*}[t!]
\small
\begin{center}
\begin{tabular}{cccc}
{\bf RGB} & {\bf SVM} (81.11\%, 0.82) & {\bf Bag kernel} (83.44\%, 0.83) & {\bf Mean map} (85.21\%, 0.84)\\ 
\includegraphics[width=3.5cm]{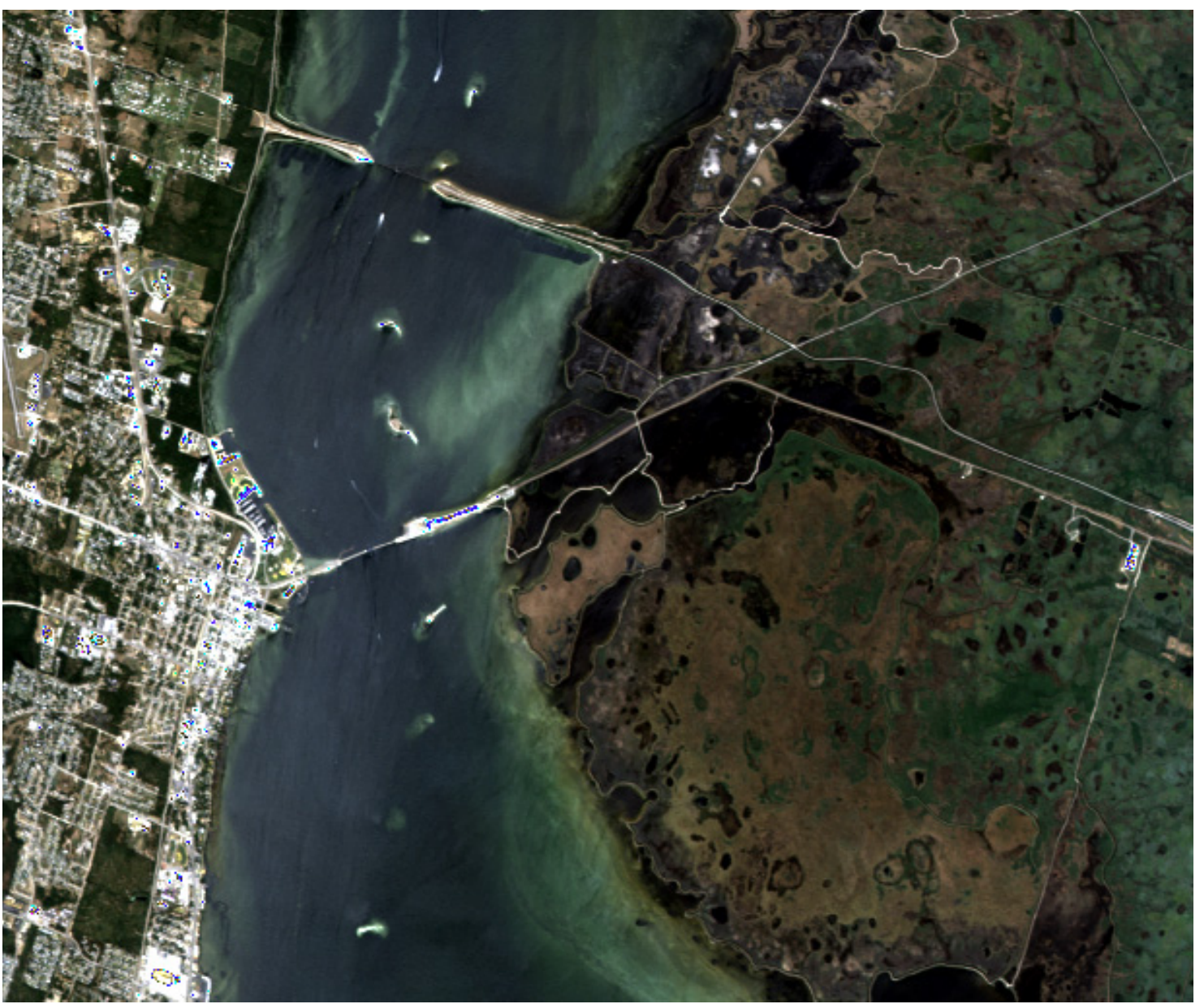} &
\includegraphics[width=3.5cm]{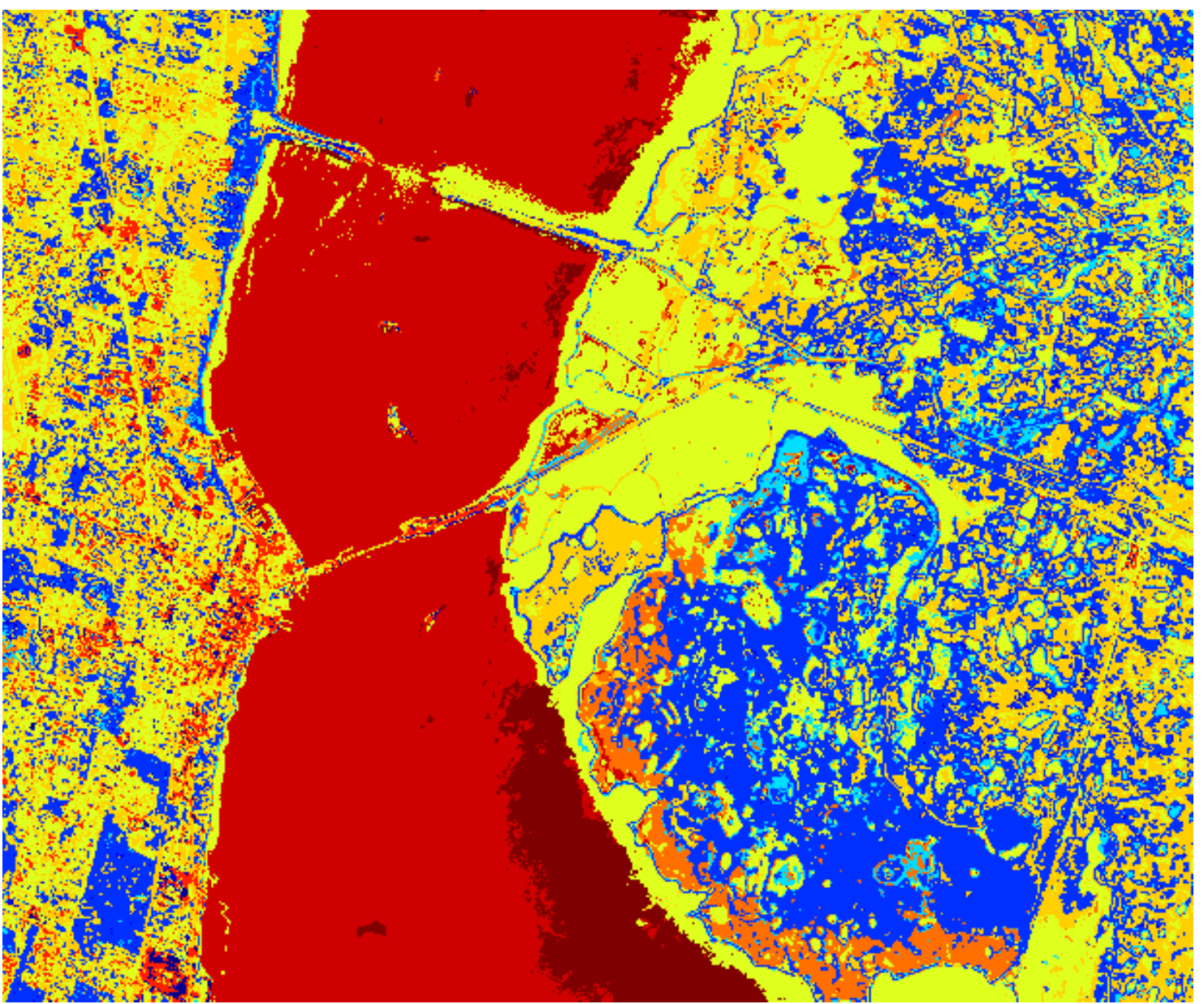} & 
\includegraphics[width=3.5cm]{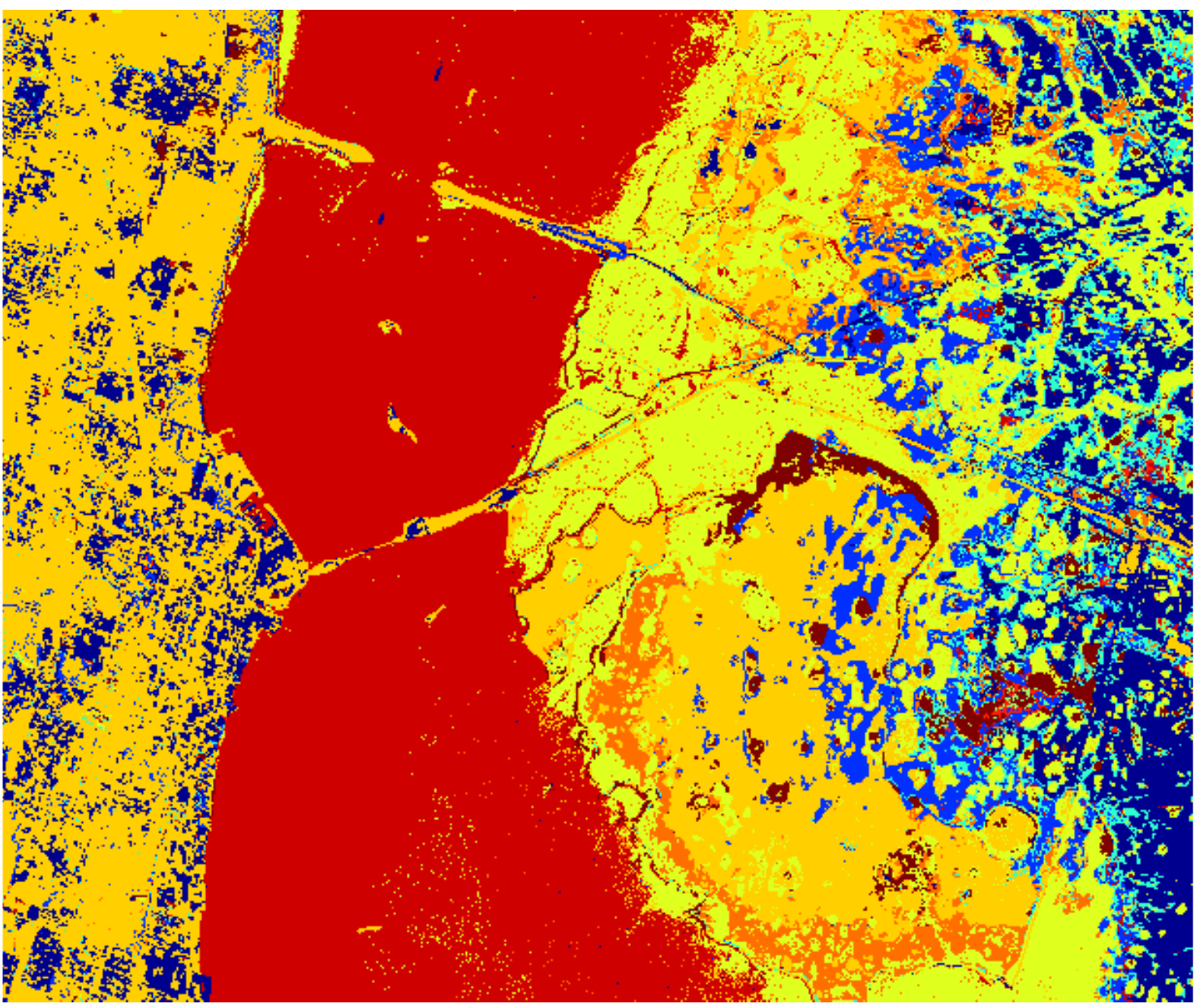} &
\includegraphics[width=3.5cm]{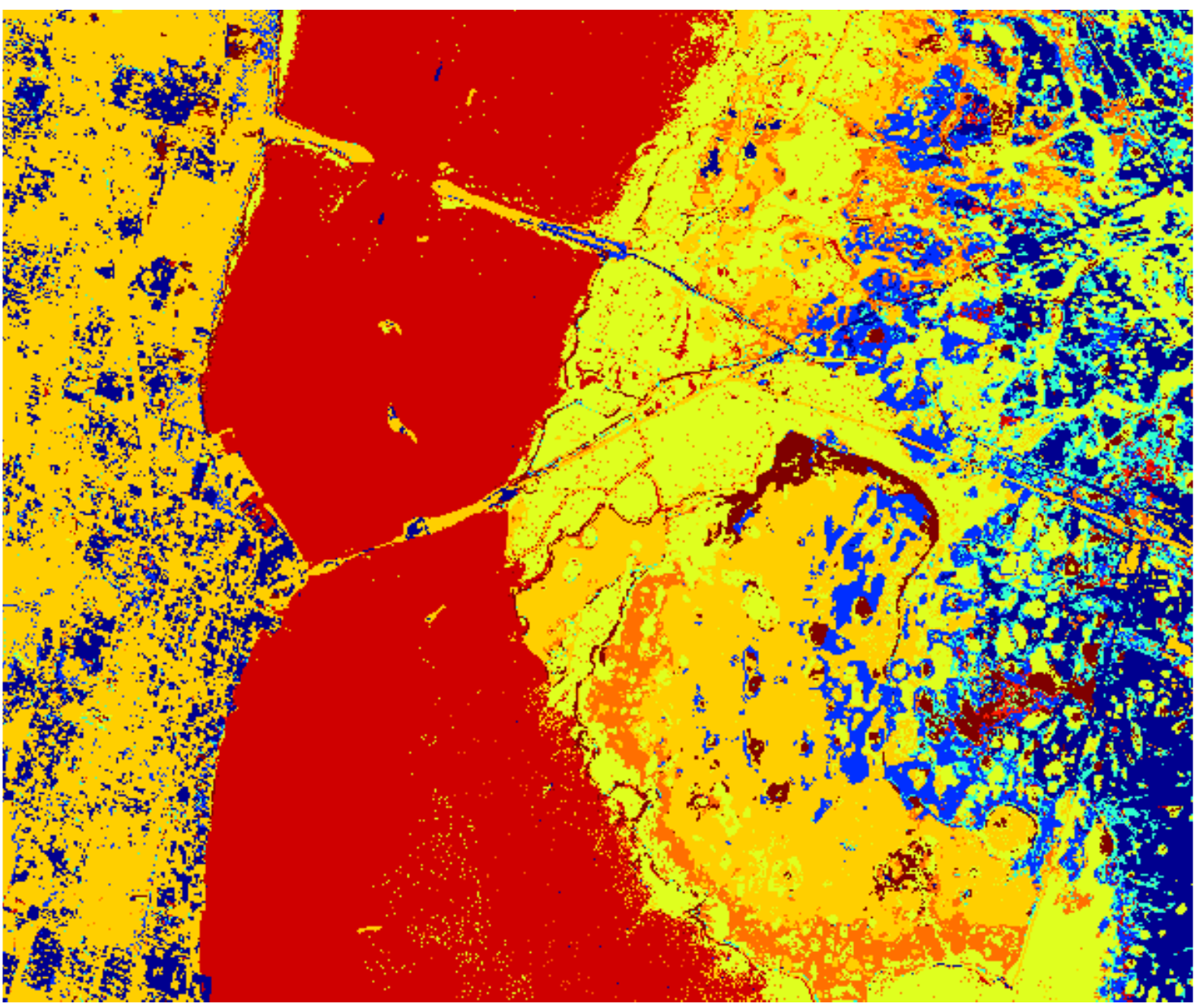}\\
& {\bf Label prop.} (70.57\%, 0.64) & 
{\bf LapSVM} (83.11\%, 0.83) & 
{\bf SSNN (87.89\%, 0.87)}\\ 
& \includegraphics[width=3.5cm]{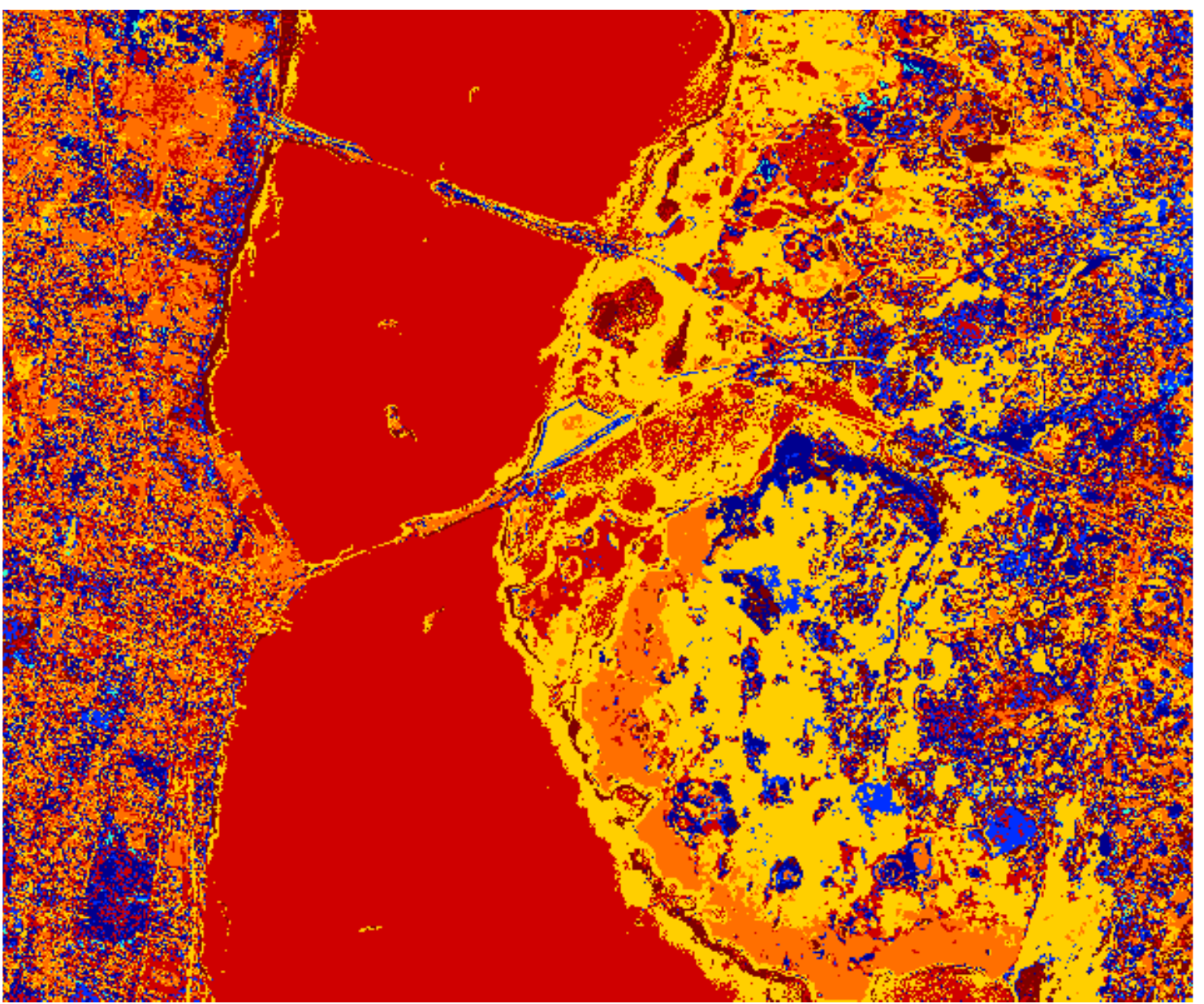} & 
\includegraphics[width=3.5cm]{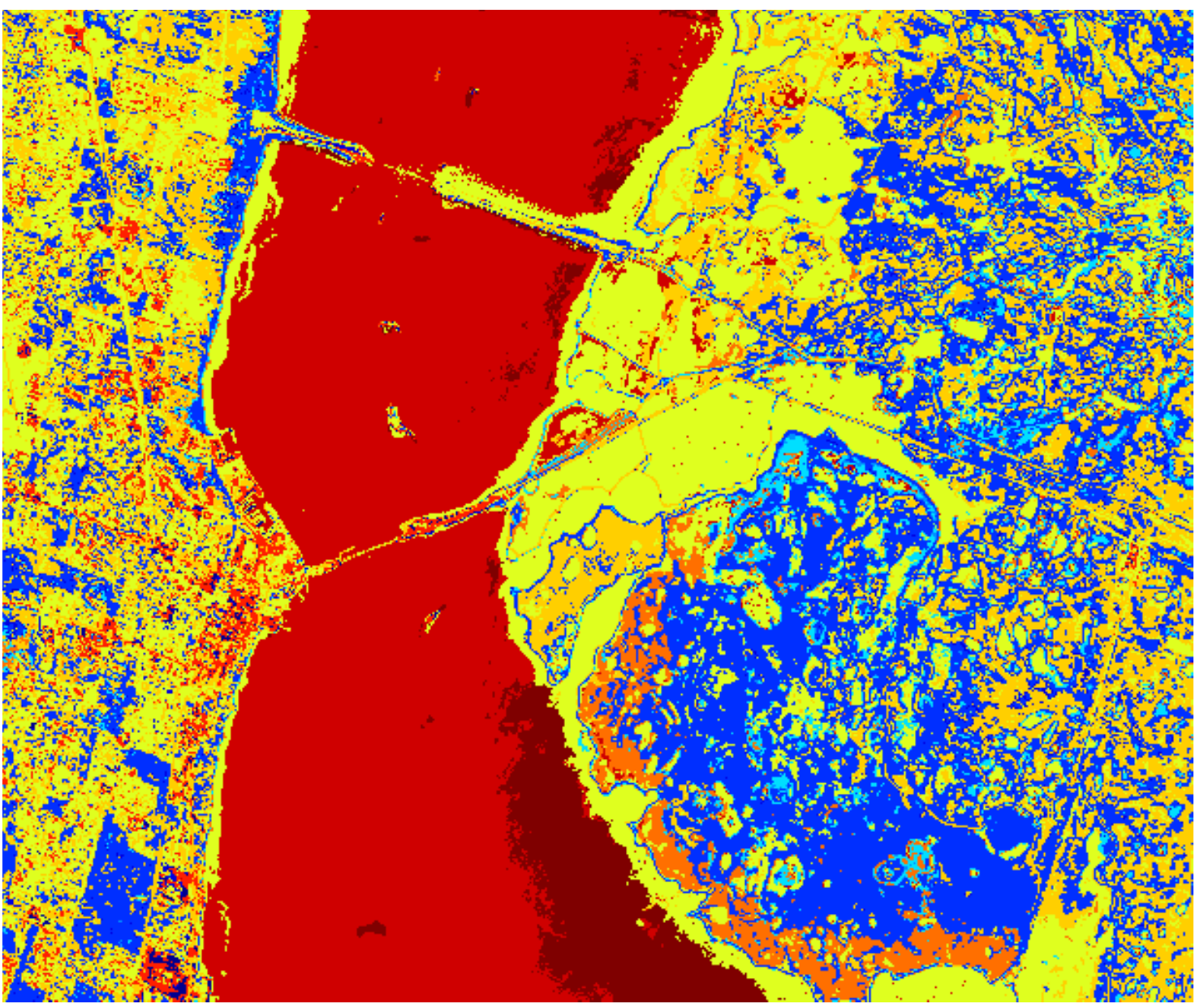} &
\includegraphics[width=3.5cm]{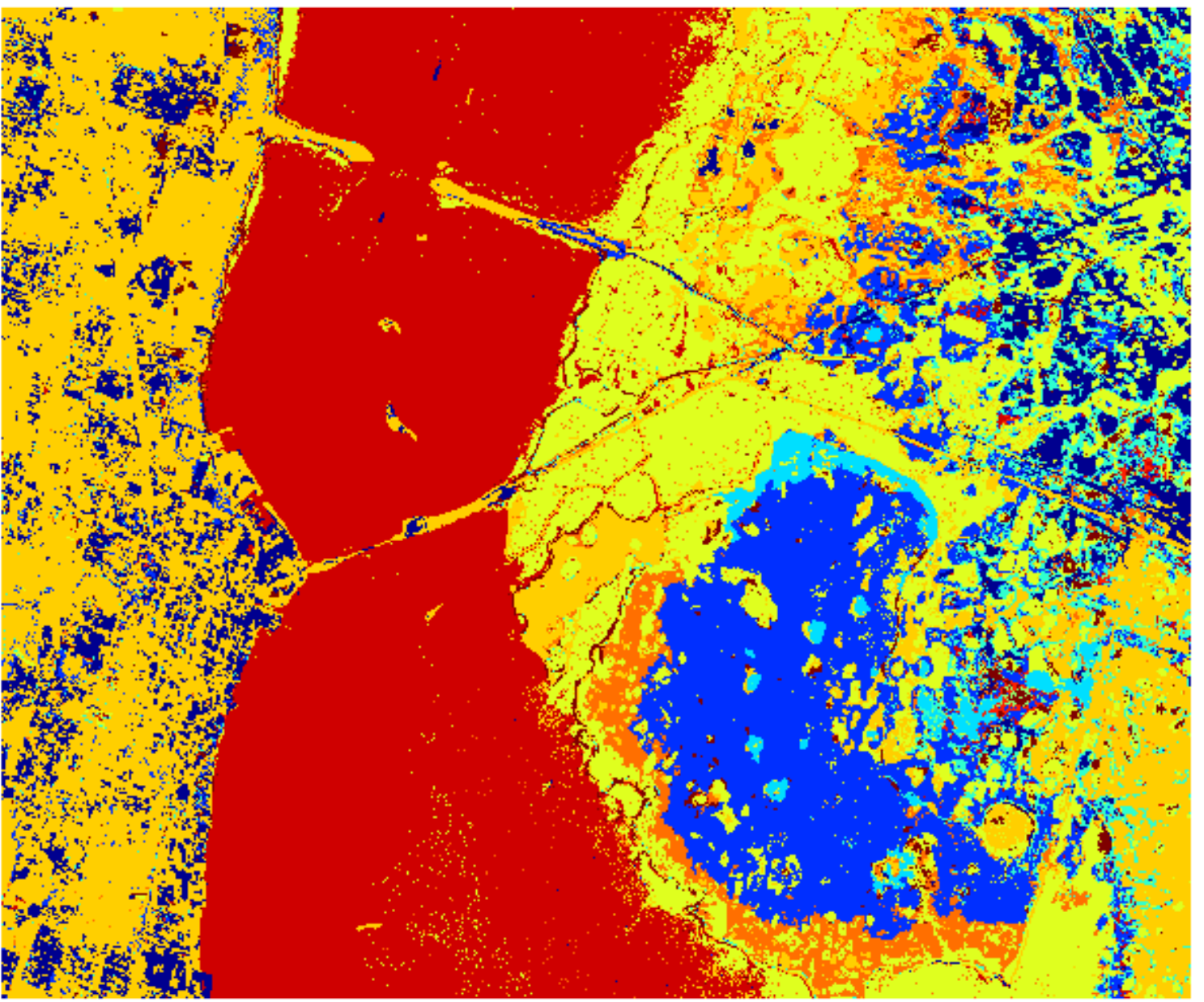}
\end{tabular}
\end{center}
\vspace{-0.5cm}\caption{RGB composition 
and  classification maps with SVM, bag kernels, probabilistic mean map kernel, 
label propagation, LapSVM, and SSNN for the KSC image ($\ell=200$, $u=1000$). Overall accuracy and kappa statistic are given in brackets.}
\label{fig:ksc-classified}
\end{figure*}

Figure~\ref{fig:ksc-classified} illustrates classification results for cluster kernels, probabilistic mean map kernel, 
label propagation, Laplacian SVM (LapSVM), and semisupervised neural networks (SSNN). We used $\ell=200$ labeled pixels ($20$ per class) and $u=1000$ unlabeled pixels. LapSVM, cluster kernels and mean map kernels perform similarly, and all improve the results of the label propagation whose training was particularly difficult in this high dimensional setting. 
More homogeneous areas and better classification maps are observed in general for the mean map and bag kernels, and particularly for the SSNN, which efficiently deals with complex marsh areas (south east area of the image) and cope with large scale datasets.

\begin{table}[!t]
\caption{{Summary of  active learning algorithms~\cite{Tuia11} ($c$: number of candidates, $p$: members of a committee of learners).}}
\label{tab:AL}
\small
\setlength{\tabcolsep}{1pt}
%
%
%
\blue{
\begin{tabular}{p{1.3cm}|p{1.3cm}| p{1.9cm}|p{1.5cm}|p{2.2cm}}
\hline
 Criterion&     Classifier& Uncertainty& Diversity& Models to train\\\hline\hline
 EQB&  $All$& Agreement of a committee& $\hspace{0.75cm}\times$ &  $p$ models\\\hline
 AMD&   $All$& Agreement of a committee&  $\hspace{0.75cm}\times$ &  $p$ models\\\hline
 MS&      SVM & Distance to SVM margin& $\hspace{0.75cm}\times$ &  1 SVM\\\hline
 cSV&    SVM&  Distance to SVM margin&  Spectral distance to current SVs& 1 SVM + distances to SVs \\\hline
 MOA&    SVM& Distance to SVM margin & Angular differences   & 1 SVM + distances to already selected samples\\\hline
 MCLU-ECBD& SVM& Distance to SVM margin & Different cluster assignment & 1 SVM + nonlinear clustering of $c$ samples \\\hline
 KL-max&    Prob. output& Divergence of PDF if adding the candidate & $\hspace{0.75cm}\times$&    $(c-1)$ models\\\hline
 BT&      Prob. output& Difference in posterior of most confident classes& $\hspace{0.75cm}\times$&    1 model\\\hline
\end{tabular}
}
\end{table}

\vspace{-0.25cm}\paragraph*{\small Regularization via user's interaction.} Another possibility to cope with small sample problems is to provide additional labeled examples. \blue{This is possible since HSI represent land surfaces, usually physically reachable or that can be displayed in an image processing software. Therefore, the new samples can be collected either by photointepretation of the images (only if the classes can be recognized on screen) or 
by organizing field campaigns.} However, since providing additional samples is costly, the samples to be labeled must be selected carefully. To this end, \emph{active learning} (AL~\cite{Tuia11,Cra12}, right panel of Fig.~\ref{fig:SSAL})  has gained popularity in the last years: rather than proceeding by random sampling or stratification (i.e. sampling according to a measure of the expected variability within a class), AL uses the outcome of the current model to rank the unlabeled pixels according to their expected importance for future labeling. The aim is to detect the most difficult (and diverse) pixels for the current classifier. The top ranked pixels are then screened by a human operator, who provides the labels, enlarging the training set. With the enlarged training set, a new improved classifier is built and the process is iterated. Since AL focuses on difficult areas, 
it boosts the performances with fewer samples than those required by random sampling. Figure~\ref{fig:AL} illustrates an example of active learning model regularization  for the specific task of adapting classifiers to multiple scenes. 

\vspace{-0.25cm}\paragraph*{\small Regularization through sparsity promotion.} 
Despite their high dimensionality, hyperspectral pixels belonging to the same class typically 
lie in a low-dimensional subspace. This observation was exploited in local 
semisupervised classifiers, and has been recently used in sparse signal representations. 
Here, the assumption is that pixels can be represented accurately 
as a linear combination of a few training samples from a structured dictionary. 

\blue{Note the connection with SVMs that embed the dictionary
(the training samples) into a high dimensional feature space ${\mathcal H}$. The use of the hinge loss} in the 
SVM functional induces a sparse solution, i.e. few training examples 
are selected. \blue{Recently, sparse kernel methods have been presented, such as the 
kernel matching pursuit, the $\ell_1$-SVM, the kernel basis pursuit or the generalized LASSO. In all these, 
the dictionary functions are the kernels centered around the selected `support vectors'. 
Alternatively, in~\cite{Chen12}, several sparse kernel approaches have been presented with a 
different philosophy: the target pixel is the test pixel itself, not a similarity evaluation, and the 
dictionary is composed by the training pixels in the feature space.} \blue{{In this paper}, a Basis Projection (BP) approach is 
used to promote sparsity with $\Omega = \|{w}\|_1$, as a relaxation of the more computationally demanding problem induced by using the $\ell_0$-norm.
To solve the BP problem, greedy algorithms, such as the orthogonal matching pursuit (OMP) and the subspace pursuit (SP) methods, can be used. }
The dictionary can be obtained off-line or from the same 
image. 
The classification can be additionally improved by incorporating the contextual information
from the neighboring pixels into the classifier (see the next Section). 

\begin{figure}[t!]
\begin{center}
\includegraphics[width=6cm]{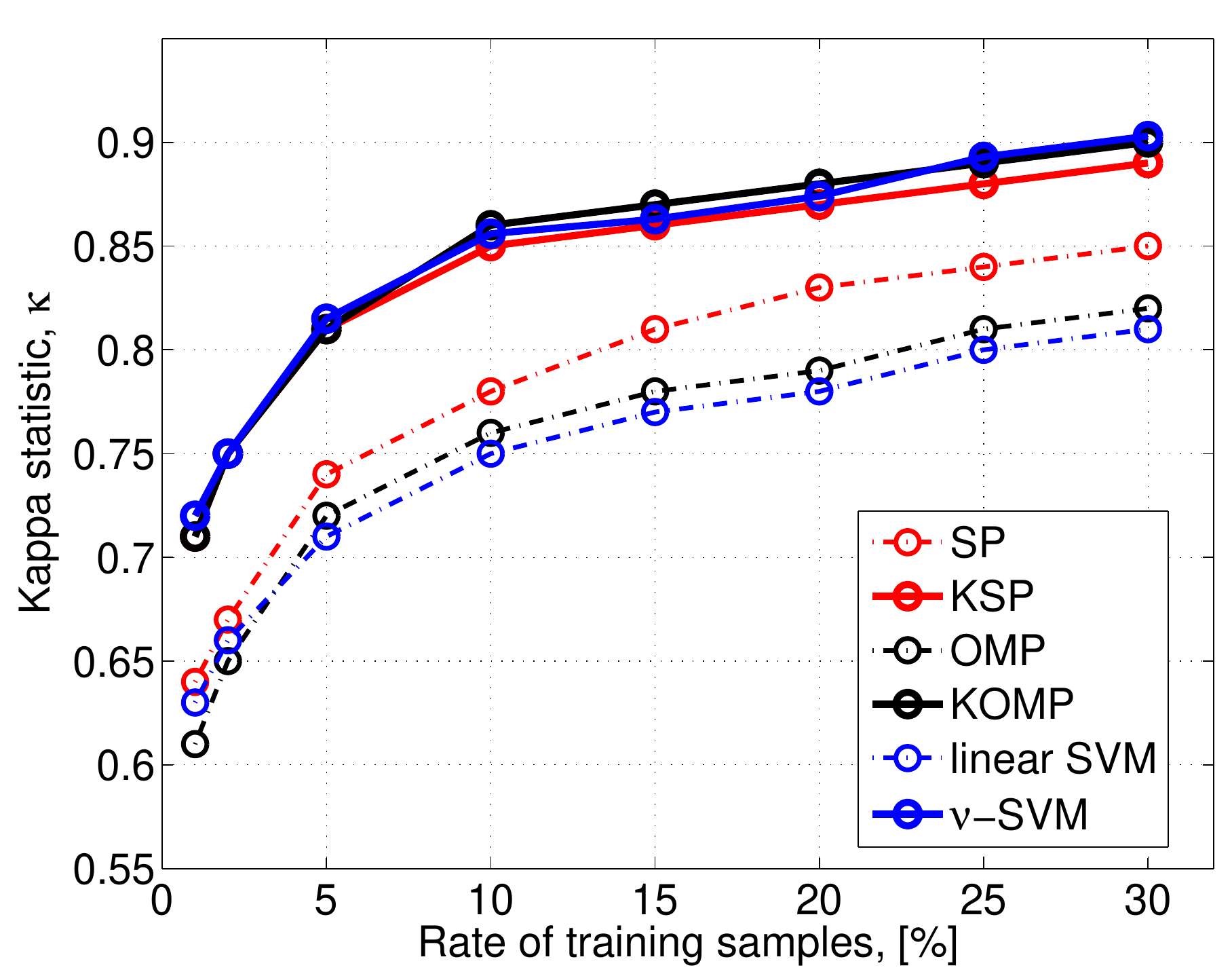}
\end{center}
\vspace{-0.5cm}\caption{Performance measure with the estimated Cohen's kappa statistic, $\kappa$, for different sparsity-promoting classifiers.}
\label{fig:sparse}
\end{figure}

In the example illustrated in Fig.~\ref{fig:sparse}, we compare the performance of the linear (SP, OMP) and kernel (KSP, KOMP) sparse HSI classifiers introduced in~\cite{Chen12}. The sparseness factor was tuned for best performance. 
We also included a linear SVM and the $\nu$-SVM using an RBF kernel, in which the parameter $\nu \in (0,1)$ controls the degree of sparsity. 
The RBF $\sigma$ parameter was tuned by standard $10$-fold cross-validation. Figure~\ref{fig:sparse} shows the 
results for these methods and different number of training samples in the standard AVIRIS Indian Pines 
hyperspectral image ($220$ spectral channels and spatial resolution $20$ m, shown in Fig.~\ref{fig:seg}). 
\blue{This is the standard benchmark hyperspectral image which is used here to allow comparison with results in~\cite{Chen12}.} 
We split the data into a training set (20\% of the available labeled pixels) and a test set (80\%). We trained the classifiers for different rates \{1, 5, 10, 15, 20, 25, 30\}\% of the training set, and show results for the test set that remained constant. 
Nonlinear methods show a much better performance 
over linear approaches. In the linear case, SP clearly outperforms the rest, but when the nonlinearity 
is included all methods perform very similarly. 

%% file: 04_spatial.tex

\section*{\small{SPATIAL-SPECTRAL IMAGE CLASSIFICATION}}\label{sec:spat}

\begin{figure*}[t!]
\small
\begin{center}
\setlength{\tabcolsep}{2pt}
\begin{tabular}{ccccccc}
RGB Composition & Reference & SVM 
& EMAP  & DBFE+EMAP  & EMAP+CSVM \cite{CampsValls06}  & EMAP+GCSVM \cite{Li13} \\
&& $\omega$ & $s$ & $s$ & $\omega + s$ & $\omega + s$\\
&& (81.01) 
& (89.89) & (94.50) & (97.80) & (98.09) \\
\includegraphics[width=2.3cm]{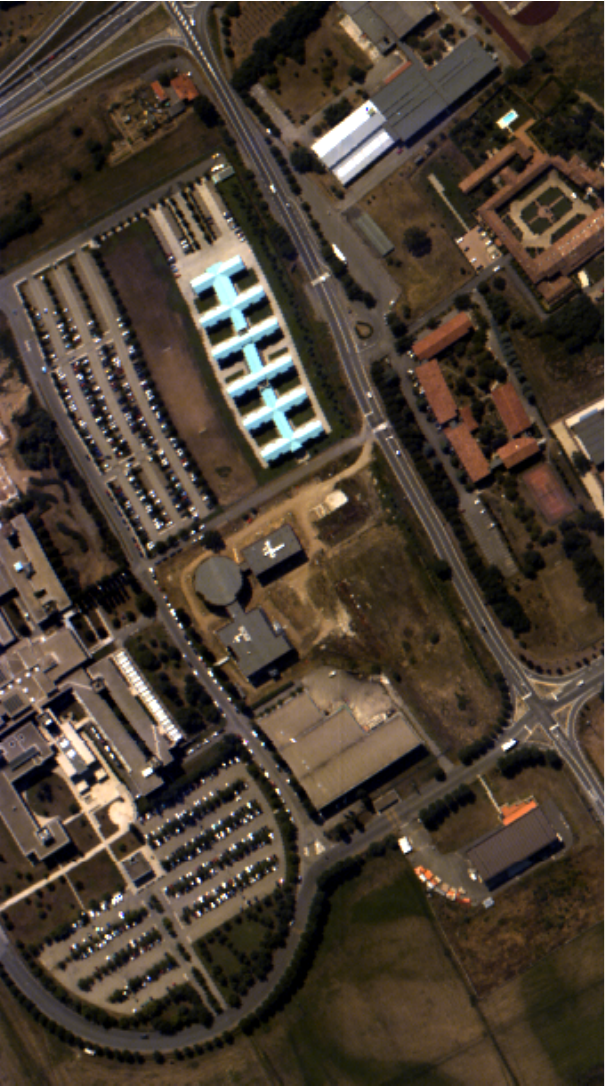}&
\includegraphics[width=2.3cm]{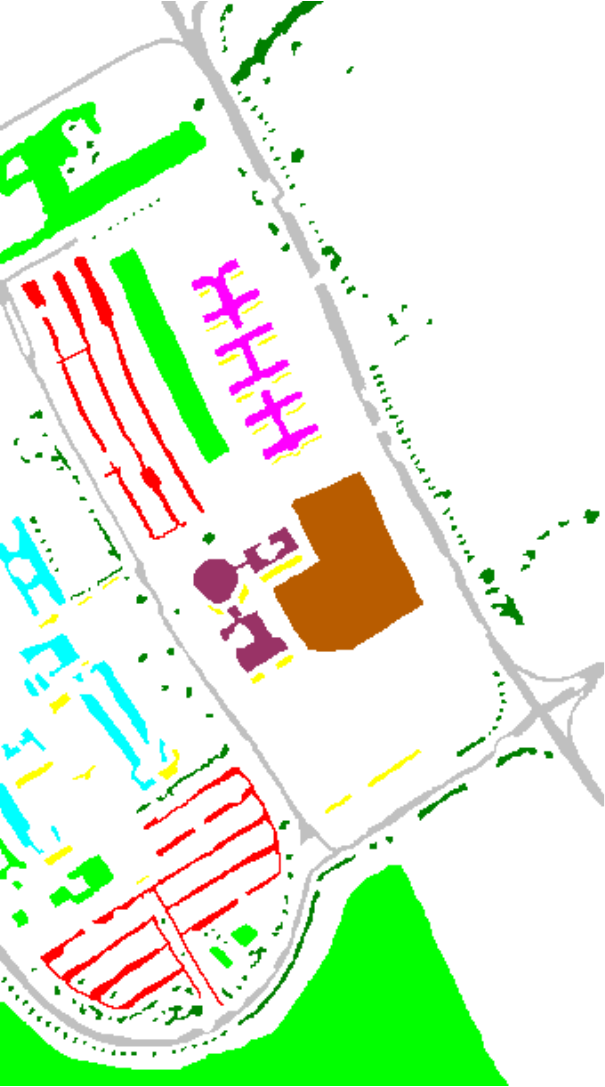}&
\includegraphics[width=2.3cm]{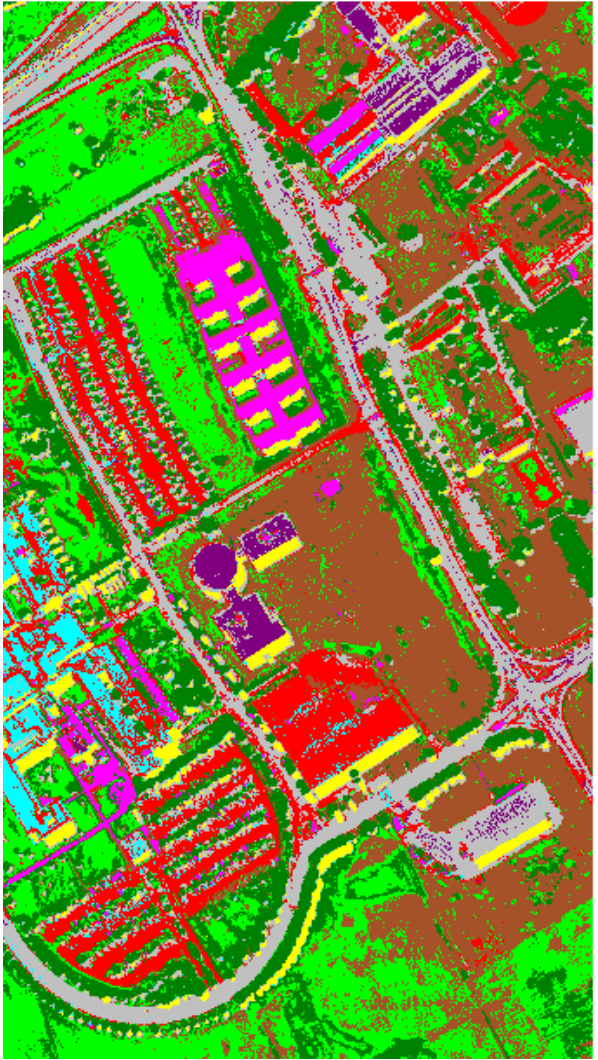}&
\includegraphics[width=2.3cm]{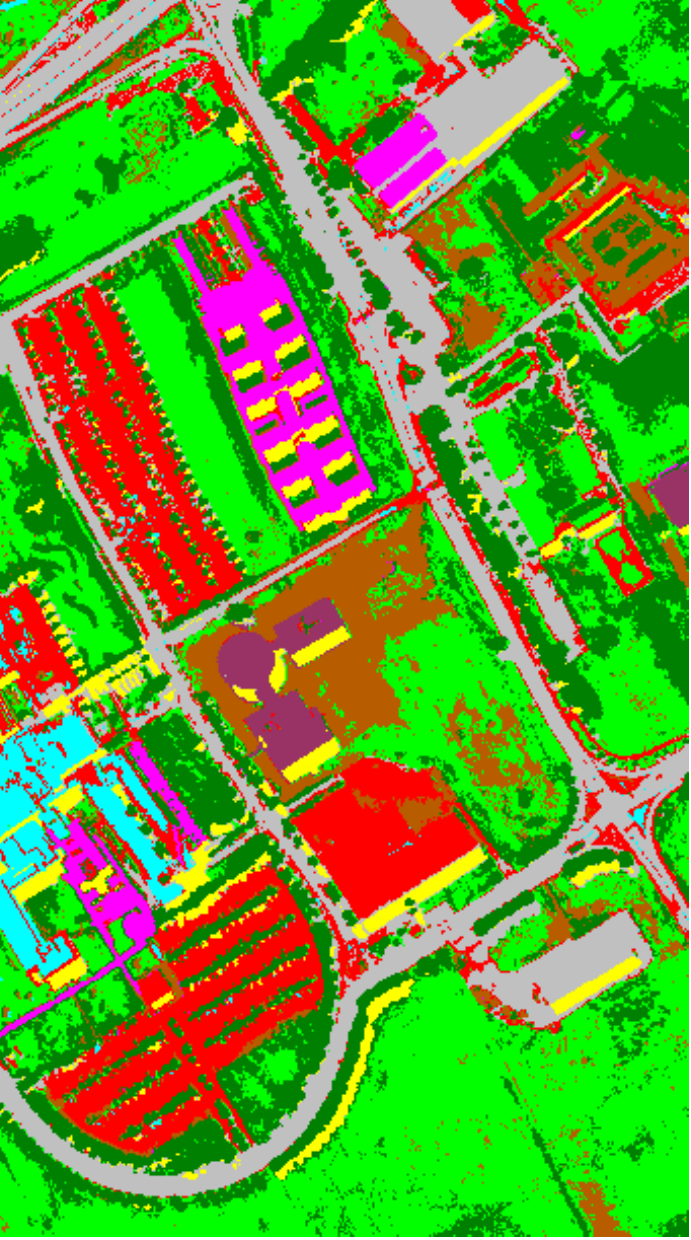}&
\includegraphics[width=2.3cm]{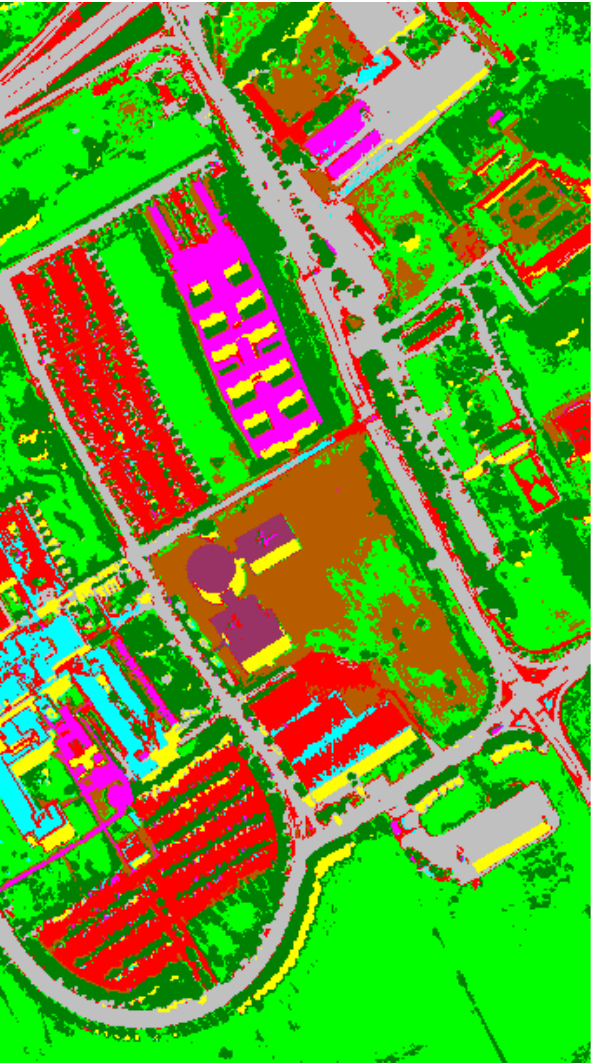}&
\includegraphics[width=2.3cm]{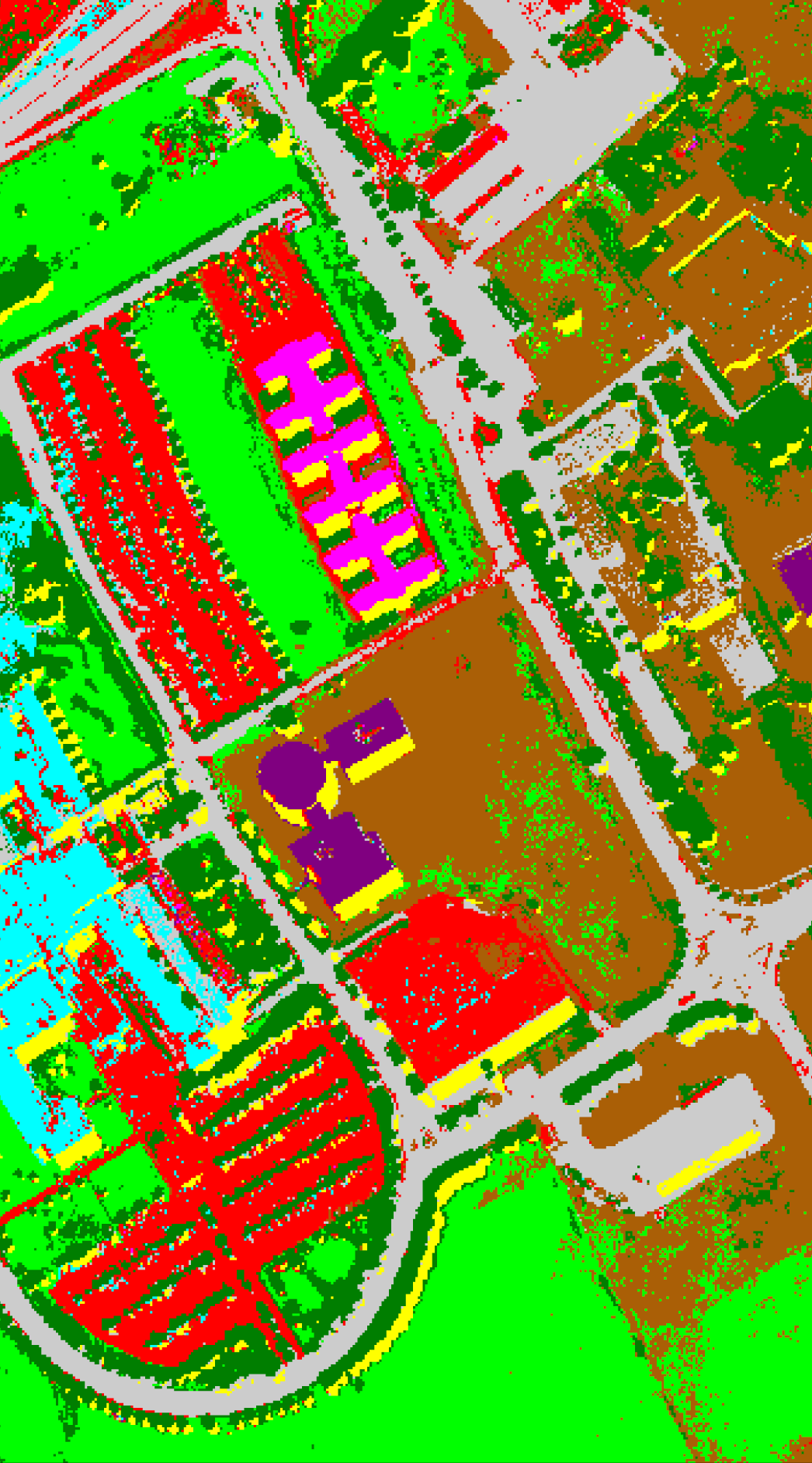}&
\includegraphics[width=2.3cm]{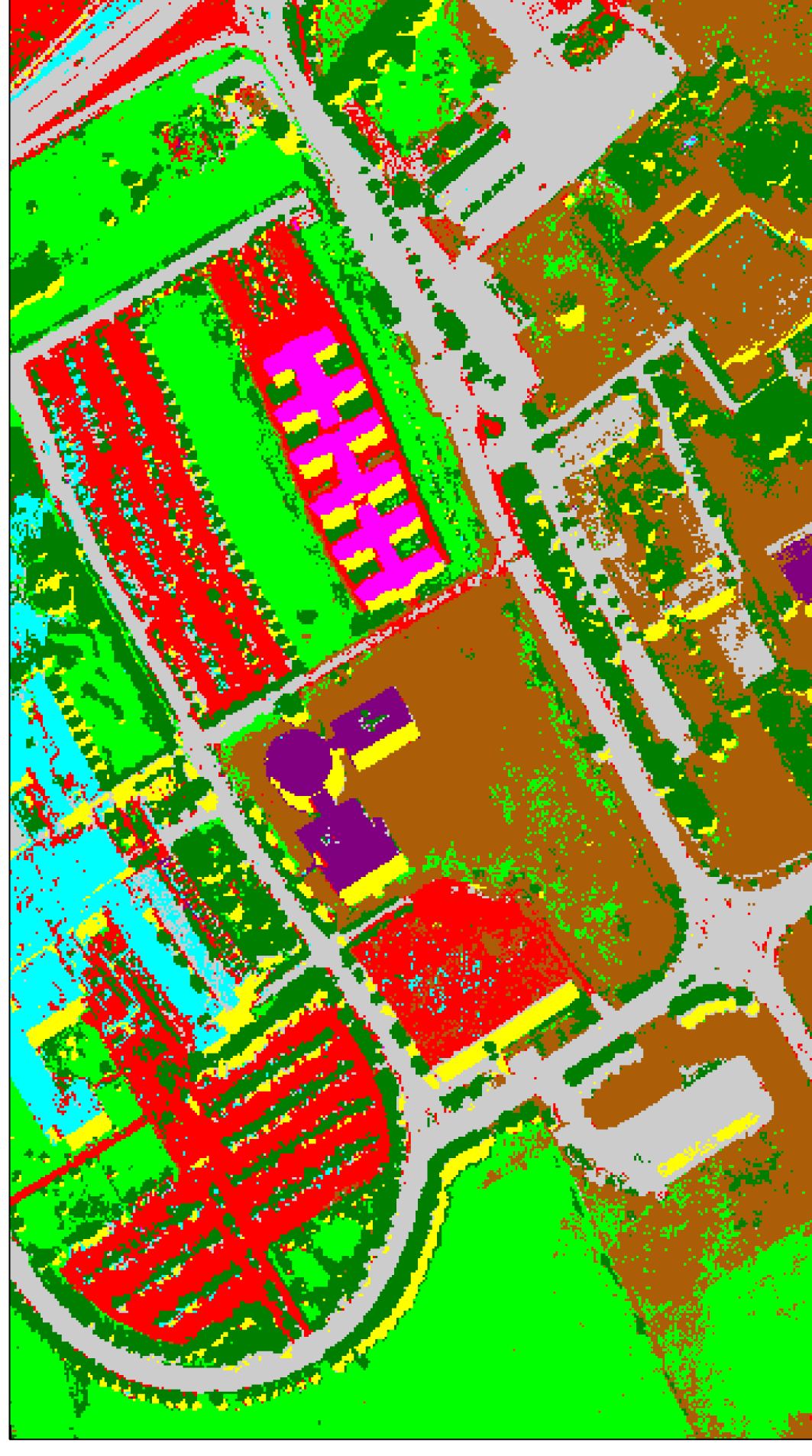}\\
\end{tabular}
\vspace{-0.5cm}
\caption{RGB composition along with the available reference data for the ROSIS-03 Pavia University area data set (103 spectral channels and spatial resolution 1.3m). Classification maps are shown for SVM on the original image only against spatio-spectral classification with the EMP, EMAP and EMAP after feature extraction by DBFE (81 data channels), composite kernels wth cross-kernels and SVM~\cite{CampsValls06}, and generalized composite kernels with multinomial logistic regression~\cite{Li13}. Overall accuracies [\%] are reported in parentheses ($\omega$: using spectral bands, $s$: using spatial filters from PCA). 
}
\label{fig:spat}
\end{center}
\end{figure*}

Hyperspectral images live in a {\em geographical manifold}, in the sense that spatially neighboring pixels carry correlated information and that images are usually smooth in the spatial domain~\cite{CampsValls11mc}[see Ch.~2]. Accounting for spatial smoothness (i) provides less salt-and-pepper  classification maps, (ii) reveals the size and shape of the structure the pixel belongs to, and (iii) allows to discriminate between structures made of the same materials, but belonging to different land-use types. Spatial regularization has been widely used to improve classification~\cite{Fauvel13}. 
The joint exploitation of both \emph{spectral} and {\em spatial} information considers that either the loss, the regularizer, or both depend on the spatial neighbourhood of a pixel~\cite{Plaza09}. 

 
\vspace{-0.25cm}\paragraph*{\small Spatial feature extraction.} 
A simple yet effective way to regularize for spatial smoothness is to enrich the input space with features accounting for the neighborhood of the pixels. This is usually done by using moving windows or adaptive filters applied to the spectral bands. These filtered images are then used to learn the classifier. 
Standard filters based on occurrence or co-occurrence, morphological operators, Gabor filters or wavelets decompositions generally provide significant improvements over purely spectral classifiers. Among them, morphological filters are the most promising. In~\cite{Benediktsson05}, filtering was performed at many scales and an \emph{Extended Morphological Profile} (EMP) was used for classification. Proceeding in a multiscale fashion enables the adaptive definition of the neighborhood of a pixel according to the structure it belongs to. Filtering in HSI is more challenging than in multispectral images, and one typically resorts to compute the EMP based on only a few Principal Components (PCs) using morphological reconstruction operators. 
 All the features are then fed to a classifier, either alone~\cite{Benediktsson05} or combined with the original spectral information~\cite{Fauvel08}. Furthermore, feature selection~\cite{Tuia10mkl} or extraction~\cite{Benediktsson05} can be used to find the relevant features. Recently, connected tree-based morphological operators have been investigated for the analysis of HSI~\cite{Mur10}. 
These so-called attribute filters extract thematic attributes of the connected components of an image which are thresholded according to their geometry (area, length, shape factors), or texture (range, entropy). The multiscale version, Extended Morphological Attribute Profile (EMAP), has been also introduced. 
In Fig.~\ref{fig:spat}, the approaches based on mathematical morphology (EMP and EMAP extracted from the first four principal components) 
are used for classification of ROSIS-03 data from an urban area in Pavia, Italy. \blue{We selected this image to illustrate the capabilities of several spatial-spectral classifiers since urban areas monitoring at VHR typically requires the extraction of directional, rotational and scale features from objects}. A significant improvement in terms of classification accuracies with respect to the spectral SVM was achieved by applying the EMAP with four different attributes on the image {(area and diagonal of the bounding box of connected components, moment of inertia and standard deviation, see~\cite{Mur10}).}  On the other hand, some redundancy was observed in the original 144--dimensional filter vector of EMAP. Therefore, Decision Boundary Feature Extraction (DBFE) was applied on it. After extraction with DBFE, the accuracies improved significantly ($+5\%$), thus confirming the importance of feature extraction routines. Finally, comparison with composite kernels (see discussion below in Sect.\ref{sect:adv}~\cite{CampsValls06,Li13}) yielded improved  classification accuracy, but with a strong change of response in the right part of the image, where much more soil is predicted. This is explained by the fact that these two approaches jointly use spectral and spatial information, and are thus closer to the original spectral SVM result (for which the right part was partially predicted as soil).

\vspace{-0.25cm}\paragraph*{\small Spatial-spectral segmentation.} 
Another approach for the inclusion of spatial information is through image \emph{segmentation}, 
typically using watershed, mean shift and hierarchical segmentation~\cite{Fauvel13}. 
After segmentation, a supervised scheme assigns the pixels in the segments to the classes. Two approaches are the mostly used: in the first, the regions are treated as input vectors in a supervised classifier. In the second regions are considered as basins to post-process the class memberships attributed by a pixel-based classifier within each segment. 

The reverse view on the problem is proposed in~\cite{Tarabalka10b}, where a supervised classifier is used to produce confidence values for each pixel. Then, pixels with maximal confidence are used as seeds for a region growing algorithm. In~\cite{Mun11} segmentation and classification are linked by user-provided labels: working with a hierarchical segmentation of the data, the labels provided are used to isolate coherent clusters, both spatially and thematically, thus ending with the good segmentation and the labels of the segments. The number of queries is minimized with active learning. 

\begin{figure}[t!]
\small
\begin{center}
\setlength{\tabcolsep}{4pt}
\begin{tabular}{cc}
RGB & SVM (78.2\%, 0.75)\\
\includegraphics[width=3.8cm]{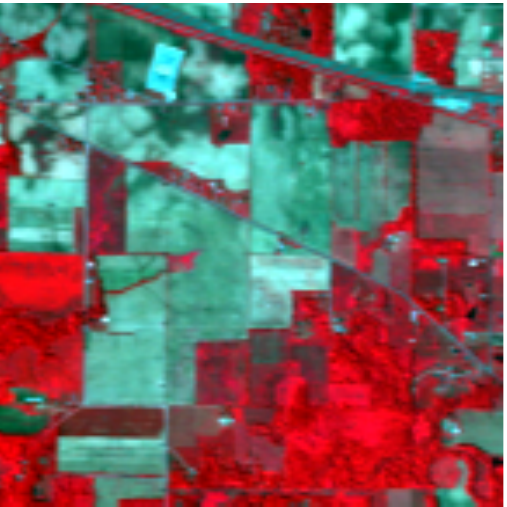}&
\includegraphics[width=3.8cm]{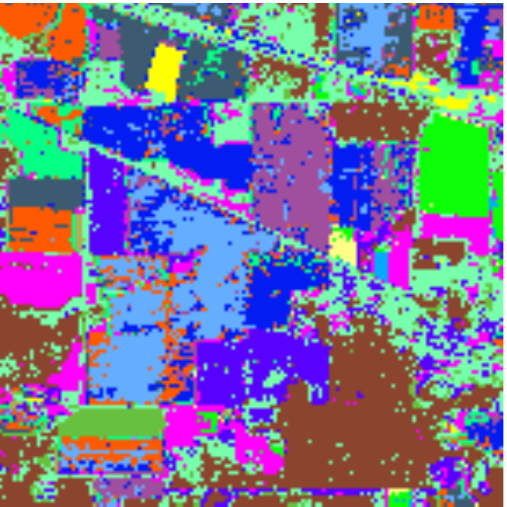}\\
Majority Voting (90.8\%, 0.90) & Markers (91.8\%, 0.91)\\
\includegraphics[width=3.8cm]{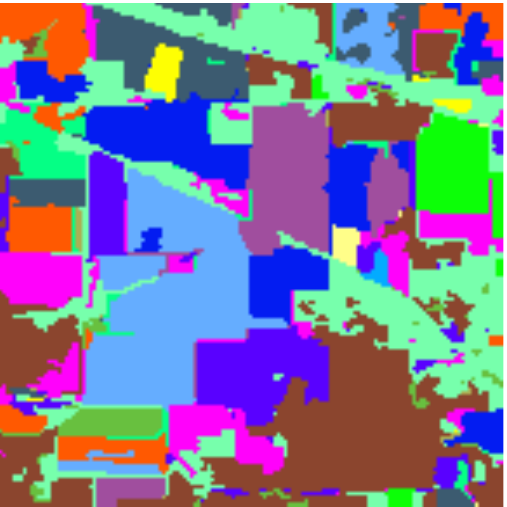}&
\includegraphics[width=3.8cm]{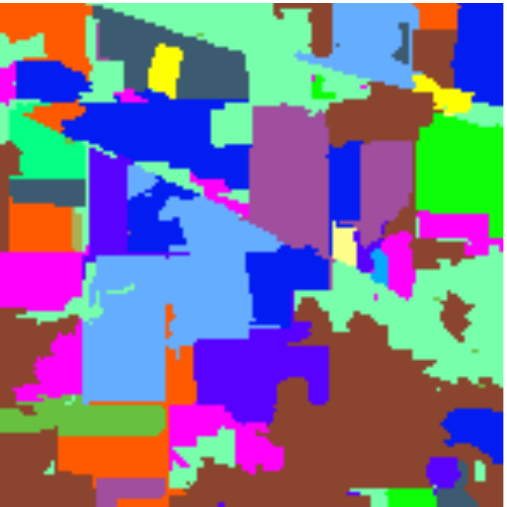}\\
\end{tabular}
\vspace{-0.3cm}
\caption{RGB composition of the standard 
AVIRIS Indian Pine data set (200 spectral channels and spatial resolution 20m). Classification maps are shown for spatio-spectral classificaiton with the segmentation and classification with majority voting and segmentation with markers against SVM on the original image only. Overall accuracies and the kappa statistic for each method are reported in parentheses~\cite{Fauvel13}. 
}
\label{fig:seg}
\end{center}
\end{figure}

In Fig.~\ref{fig:seg}, two approaches based on segmentation and classification with majority voting and markers are applied to a 220-bands AVIRIS dataset over Indian Pines (Indiana, USA). Significant improvement in terms of overall classification accuracies and kappa statistic were achieved over the pixel-based SVM classifier. Using markers provided the best accuracies: a $+1\%$ improvement over the simple majority voting and more than $+13\%$ over the traditional pixel-based SVM classifier. Furthermore, as seen in Fig.~\ref{fig:seg}, it is clear that the classification and segmentation approaches provide a significantly more uniform classification map when compared to the purely spectral SVM classification map.

\vspace{-0.25cm}\paragraph*{\small Advanced spatial-spectral classifiers.} \label{sect:adv}
The main problem with spatial-spectral feature extraction approaches is the possibly high-dimensionality of the feature vectors to feed the classifier. This was alleviated in~\cite{CampsValls06} where dedicated kernels for the spectral and spatial information were combined. The framework has been recently extended to deal with convex combinations of kernels through multiple-kernel learning~\cite{Tuia10mkl} and generalized composite kernels~\cite{Li13}. In both cases, however, the methodology still relies on performing an {\em ad hoc} spatial feature extraction before kernel computation. Other alternatives in the literature considered 
the definition of graph kernels that capture multiscale higher-order relations in a neighborhood without computing them explicitly~\cite{CampsValls10b}, and the modification of the SVM to seek for the spatial filter that maximizes the margin~\cite{Flamary12}. 

A final alternative is to include contextual information with Markov Random Fields (MRF), which naturally include a spatial term on class smoothness in the energy function. 
However, in the high dimensional context of HSI, the standard application of the neighbor system definition makes the problem computationally intractable, 
and therefore recent works have focused on joining MRF spatial priors and discriminative models in HSI classification~\cite{Tarabalka10,Moser13}. An excellent review of MRF spatial-spectral methods can be found in~\cite{Mos13}.

\begin{table}[!t]
\caption{Summary of  spatial-spectral algorithms}
\label{tab:spat-spec}
\small
\setlength{\tabcolsep}{2pt}

\begin{tabular}{p{1.7cm}|p{2.5cm}|p{3.89cm}}
\hline
Type of \white{AAA} Approach&Model &  Idea \\\hline\hline
Spatial filters extraction &
Co-occurrence & Extract texture based on statistics of pairs of pixels in a neighborhood\\
\cline{2-3}
&EMP& Multiscale mathematical morphology (based on size)\\
\cline{2-3}
&EMAP& Multiscale  mathematical morphology (variety of attribute types) \\
\hline
Spatial-spectral segmentation & Segmentation and classification based on majority voting &  All pixels are assigned to the most frequent class inside a segmented region \\
\cline{2-3}
&Segmentation and classification based on markers & Most reliably classified pixels are selected as ``region markers'' for segmentation\\
\cline{2-3}
&Semi-supervised hierarchical clustering tree & Returns both classification and confidence maps. Active learning used to select informative samples.\\
\hline
Advanced spatial-spectral  &Composite and multiple kernels & Balances between spatial and spectral information with dedicated kernels \\
\cline{2-3}
classification&Graph kernels &Takes into account higher order relations in each pixel neighborhood \\
\cline{2-3}
&MRF&  Markov Random Field Modeling (probabilistic)  \\

\hline
\end{tabular}

\end{table}

%% file: 05_adaptation.tex

\section*{\small ADAPTATION AND INVARIANCES}

One of the greatest challenges of modern HSI classification is the adaptation of classifiers between acquisitions that differ either by the zone they represent and/or the acquisition conditions such as illumination, angle and season, among other effects. Adaptation is a central issue in HSI classification. For example, the increase in revisit time of recent satellites 	
has improved multitemporal analysis of scenes. 
Nevertheless, algorithms must be able to adapt to changing situations. Generally, the direct application of classifiers trained on one image to new images leads to poor results: even if the objects represented in the images are roughly the same, differences 
in acquisition induce significant {\em local} changes in the probability distribution function (PDF). 
These changes must be modeled and introduced in the classifiers. The concept of {\em adaptation} can be implemented at the levels of image pre-processing, robust and invariant feature extraction, or in the design of the classification algorithm. 

\paragraph*{\small Preprocessing.} 

The pre-processing phase can address adaptation through the use of radiometric correction techniques applied to the images. {\em Absolute corrections} aim at transforming the radiance measured at the sensor into surface reflectance. {\em Relative calibration} techniques adapt the radiometric properties between portions of an image or between images. Generally, absolute correction techniques require additional ground reference data that in many cases are not available or are difficult to collect. Relative calibration methods are often considered as a pragmatic alternative for adaptation. Among these methods, we recall histogram matching, relative radiometric normalization of time series~\cite{Canty04}, and multivariate histogram matching~\cite{Inamdar08}. 

Images can be casted as point clouds in a geometrical space endorsed with an appropriate distance measure. Such a view is quite convenient because it allows us to move from {\em image adaptation} to {\em manifold adaptation}. Recent methods have explicitly considered the distortions occurring between image manifolds. In~\cite{Pet11}, multitemporal sequences for each pixel were aligned based on a measure of similarity between sequences barycenters, thus consisting into a global measure of alignment. In \cite{Jun11}, spectra of the pixels are spatially detrended using Gaussian processes in order to avoid shifts related to geometrical differences or to localized class variability. A recent principled approach tries to match graphs representing the data manifolds~\cite{Tuia12}. There, the graphs of the two domains are matched using a procedure aiming at maximizing their similarity, while at the same time preserving the original structure of the graphs.

\paragraph*{\small Adapting the classifier.} 
\label{sec:adaptC}
Learning a transformation between domains may be insufficient to handle all the perturbing factors, so alternative approaches are concerned with the adaptation of the classifier itself. From a machine learning perspective, the problem of classifier adaptation is studied in the framework of transfer learning, and in particular of domain adaptation. Domain adaptation reduces to learning from data in a source domain ${\mathcal S}$ (e.g. a portion of an image) to extrapolate to a different target domain ${\mathcal T}$ (another portion of the image or to another image). The problem has been payed attention in HSI classification lately~\cite{Bruzzone10}. In this setting, source and target domains are assumed to share the same set of information classes (exceptions to this constraint in~\cite{Tui11d,Jun13}) and to follow similar (but not the same) class distributions. Domain adaptation problems in remote sensing have been mainly addressed with semisupervised techniques, which exploit the labeled samples from ${\mathcal S}$ and the unlabeled samples from ${\mathcal T}$ in order to derive a classification rule suitable for the target domain. The most recent developments in this sense consider 
semisupervised and domain adaptation SVMs~\cite{Bruzzone10}, Gaussian processes~\cite{Jun11,Jun13} and the mean-map kernel methods~\cite{GomezChova10}. 

\begin{figure}[t!]
\centerline{\includegraphics[width=8.8cm]{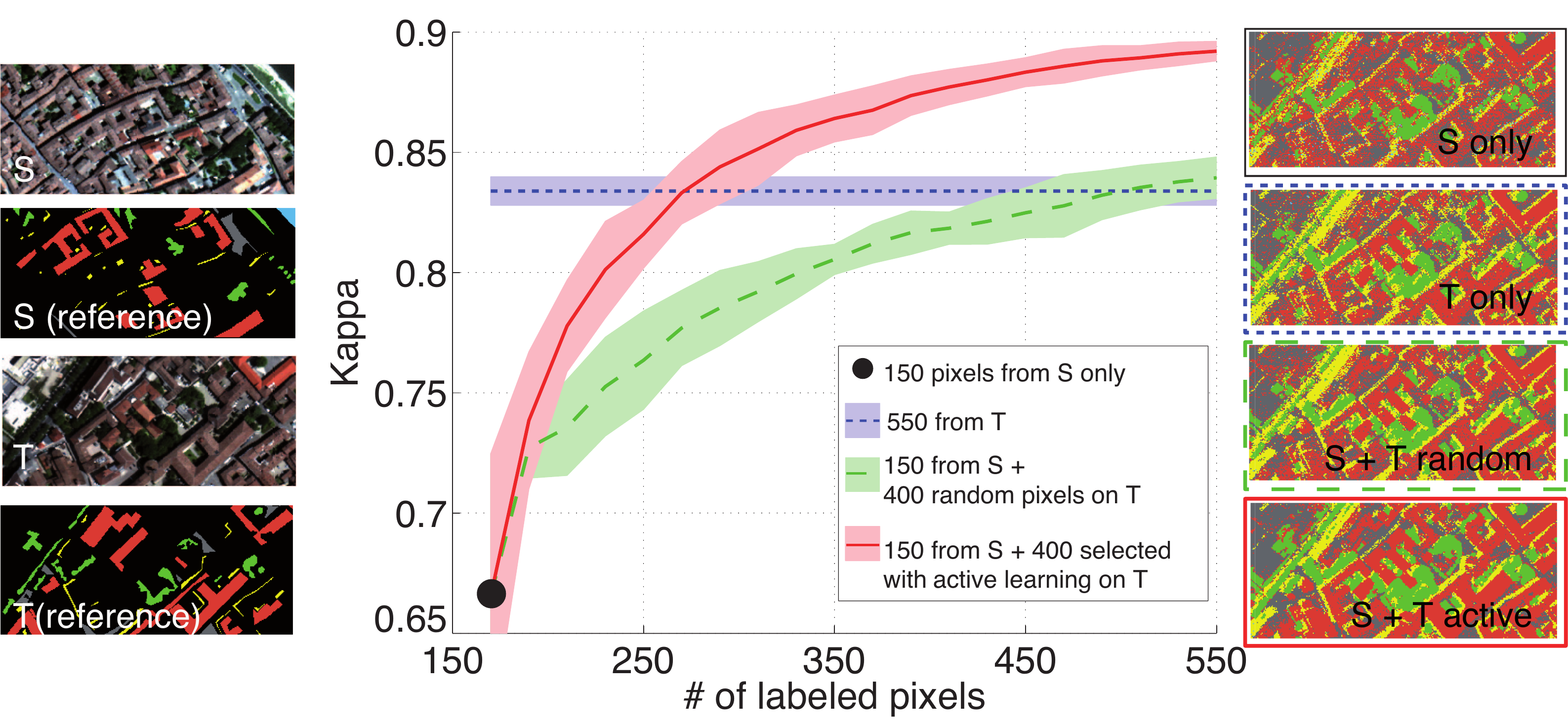}}
\vspace{-0.25cm}
\caption{Use of active learning to adapt a maximum likelihood classifier~\cite{Tui11d}. Left: given an image and a set of reference pixels in a first source area (${\mathcal S}$), we want to classify another spatially disconnected target area (${\mathcal T}$), by adding labels chosen actively in the reference of ${\mathcal T}$. Center: learning curves for the active (solid red line) and random sampling (dashed green line), evolving between the extremes of a model without adaptation (black dot at 150 samples), and another actively adapted that uses 550 samples randomly selected from ${\mathcal T}$ (blue dashed line). Right: classification maps in the target domain ${\mathcal T}$. The color of the bounding box in each map refers to the legend in the central plot.}
\label{fig:AL}
\end{figure}

Recently, active learning has been also used for adaptation assuming that some samples (as few as possible) from the target domain can be labeled by the user and added to the existing training set (defined on ${\mathcal S}$) in order to adapt the classifier to the target domain ${\mathcal S}$~\cite{Tui11d,Per12}. 
This makes the adaptation process more robust than in the case of semi-supervised learning at the cost of requiring  additional labeled samples. {Figure~\ref{fig:AL} illustrates this principle for a $102$-bands image of the city center of Pavia acquired by the ROSIS-03 airborne sensor. The whole image (only a portion is shown) has a size of $1400 \times 512$ pixels and spatial resolution is  $1.3$ m. Five classes of interest (buildings, roads, water, vegetation and shadows) are considered, and a total of $206,009$ labeled pixels are available. We explore the potential of migrating a classifier built on a source area $\mathcal{S}$ with as few labeled pixels as possible from the rest of the image $\mathcal{T}$. First, the model trained with $150$ labeled pixels randomly drawn from ${\mathcal S}$ was directly applied to ${\mathcal T}$ and yielded a $\kappa_{\mathcal S}$ of $0.67$. Noticeably, a classifier built on $550$ samples randomly selected on ${\mathcal T}$ reached $\kappa_{\mathcal S} = 0.84$. This suggests that different regions of the same scene follow very different statistics for classification. In order to improve the first classifier, we enlarged the first training set with labeled samples drawn from ${\mathcal T}$, either taken through random sampling (RS, green dashed line) or with active learning (AL, red solid line). Using AL allows to concentrate efforts in areas where the first model is suboptimal, so performance is improved with respect to RS. After $400$ queries (thus, a model using $550$ training samples in total), random sampling yields similar performance to a model using $550$ randomly drawn pixels ($\kappa_{\text{RS}}=0.84$), while active learning improves the results with $\kappa_{\text{AL}}=0.89$. \blue{To reach the performance of the model using 550 random pixels, AL requires only 120 active queries (thus a total of 270 samples in the model)}.}

\paragraph*{\small Extracting and encoding invariances in the classifier.}

\blue{Image classifiers must be robust to changes in the data representation within each land cover class}. The property of such mathematical functions is called `invariance'. 
A classifier should be invariant to object rotations, to changes in illumination, to the presence of shadows, and to the spatial scale of the objects to be detected. 
Extracting robust features (invariants) for classification and domain adaptation has been traditionally pursued by looking at the spatial or the spectral signal characteristics. On the one hand, {\em scale invariants} aim to make classifiers invariant to perturbations of object scales. In HSI classification, a single spatial scale is typically suboptimal because different classes exhibit diverse sizes, shapes, and internal variations. Multiscale classification schemes may alleviate these problems. Also, following wavelet-based representations and exploiting SIFT descriptors, {\em translation invariants} have been recently explored. 
On the other hand, {\em spectral invariants} are considered the fundamental descriptors of object structure, and are commonly employed to characterize canopy structure. 
Spectral invariance to daylight illumination allows for example improving multitemporal image classification. 

Incorporating invariances in SVM can be achieved by designing particular kernel functions that encode local invariance under transformations, or to generate artificial examples for training to which the model must be invariant. \blue{In the following example, we consider the latter possibility, with the Virtual SVM (VSVM) method}, which has been successfully exploited to encode scale, rotation, translation and shadow invariance in HSI classification~\cite{Izquierdo13}. 

\begin{figure}[!t]
\small
\begin{center}
\setlength{\tabcolsep}{1pt}
\begin{tabular}{ccc}
RGB & SVM (0.79$\pm$0.11) & {\bf VSVM (0.84$\pm$0.09)}\\
\includegraphics[width=2.8cm]{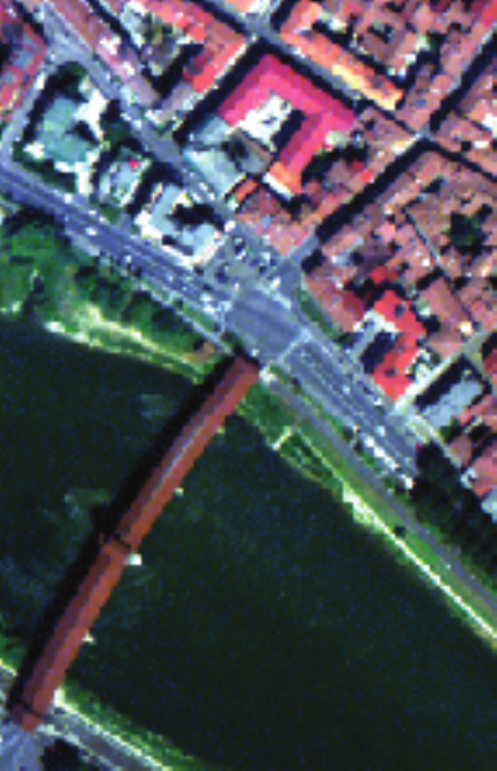}&
\includegraphics[width=2.8cm]{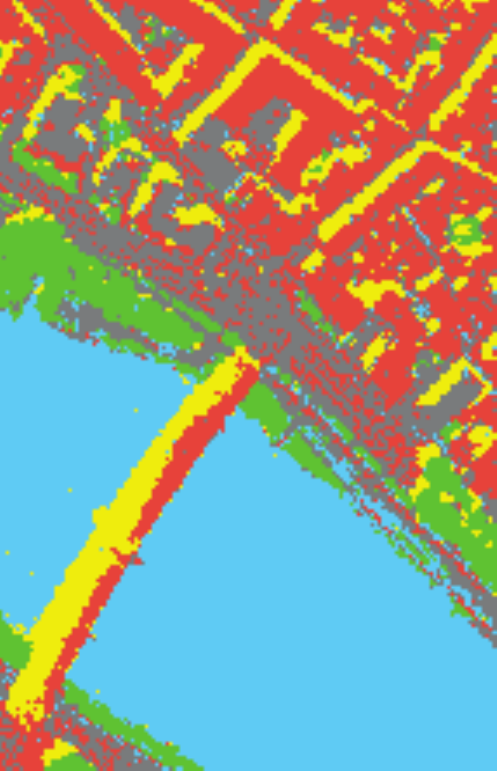}&
\includegraphics[width=2.8cm]{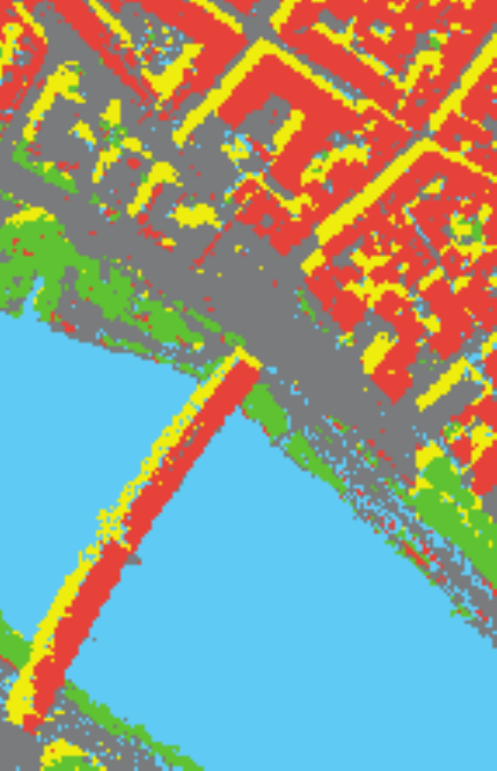}\\
\end{tabular}
\end{center}
\vspace{-0.15cm}
\caption{Experiment of patch-based classification with the virtual SVM encoding shadow invariance. True-color composite (left), and classification maps using the standard SVM (middle) and the VSVM (right) obtained using $50$ training pixels. Results are shown in parentheses in the form of (mean$\pm$ standard deviation of $\kappa$ in 20 realizations).\label{results2}}
\end{figure}

In Fig.~\ref{results2}, we illustrate the use of the VSVM encoding (spectral) shadow-invariance for image classification. We use the same data acquired by the ROSIS-03 optical sensor of the city center of Pavia (Italy) used in Fig.~\ref{fig:AL}. 
We perform patch-based classification using only $50$ training patches of size $w=5$. The classes to be detected are, as for Fig.~\ref{fig:AL}, buildings, roads, water, vegetation and shadows. 
Virtual support vectors (VSVs) were generated  according to the observed  exponential behavior of the ratio shadow/sunlit as a function of the wavelength~\cite{Izquierdo13}. 
\blue{Numerical results, as well as the zoom on a detail of the classification map, show that VSVM leads to more accurate results than the standard SVM:  encoding shadow invariance reduces misclassifications on the bridge area and an overall more homogeneous classification over flat areas (see for example the crossroads in the center of the image)}. 

%% file: 06_conclusions.tex

\section*{\small CONCLUSIONS AND DISCUSSION}

This paper reviewed and analyzed the recent developments in hyperspectral image classification. Even though HSI follow similar spatial, spectral and spatial-spectral image statistics to those conveyed by conventional photographic images, the hyperspectral signals impose additional challenges related to their high dimensionality and heterogeneity. Therefore, even though standard techniques in image processing and computer vision may be transported drectly, HSI impose important constraints to develop efficient and effective classifiers.  

The use of methods derived from statistical learning theory (SLT) has been a driving factor in recent years. SLT constitutes a proper framework to tackle the problems posed by hyperspectral remote sensing images, which typically involve scenarios with high dimensional data and few training samples. SLT permit to embed numerical regularization in nonlinear classifiers, and also to design alternative forms of conditioning and incorporation of prior knowledge.  Additionally, classification is often improved by including spatially-based and manifold-based regularizers. \blue{SLT also allows to design sparse methods able to work in relevant feature subspaces, where compact and computationally efficient method can be run}. Finally, the SLT framework has allowed to include prior knowledge in a very fruitful way: for example, classifiers can now incorporate spatial and spectral invariances that disentangle ambiguities present in land cover classification.

The field is moving fast, and attracts research from the computer vision and machine learning communities. \blue{New approaches are introduced regularly and permit to tackle new scenarios issued from high resolution imaging (e.g. multitemporal, multiangular), while learning the relevant features via robust classifiers. It should be also noted that, with upcoming satellites, efficient algorithms for dimensionality reduction before classification and fast/parallel computing solutions will be necessary to accelerate the interpretation and efficient exploitation of hyperspectral images.}